\documentclass[10pt,twocolumn,letterpaper]{article}

\usepackage[pagenumbers]{cvpr} %

\usepackage[dvipsnames]{xcolor}

\definecolor{cvprblue}{rgb}{0.21,0.49,0.74}
\usepackage[pagebackref,breaklinks,colorlinks,citecolor=cvprblue]{hyperref}

\def\paperID{263} %
\def\confName{3DV\xspace}
\def\confYear{2025\xspace}

\title{EgoGaussian: Dynamic Scene Understanding from Egocentric Video with 3D Gaussian Splatting}

\author{%
Daiwei Zhang$^{1}$ \quad 
Gengyan Li$^{1,2}$ \quad 
Jiajie Li$^{1}$ \quad 
Mickaël Bressieux$^{1}$ \quad \\
Otmar Hilliges$^{1}$ \quad
Marc Pollefeys$^{1,3}$ \quad
Luc Van Gool$^{1,4,5}$ \quad
Xi Wang$^{1}$ \quad \\[6pt]
$^1$ETH Zürich \quad 
$^2$Google \quad 
$^3$Microsoft \quad 
$^4$KU Leuven \quad
$^5$INSAIT, Sofia \\
}

\newcommand{\methodname}{EgoGaussian\xspace}
\newcommand{\myparagraph}[1]{\noindent\textbf{#1.}}

\newcommand{\update}[1]{\textcolor{black}{#1}}

\begin{document}

\twocolumn[{%
\renewcommand\twocolumn[1][]{#1}%
\maketitle
\begin{center}
    \centering
    \captionsetup{type=figure}
    \includegraphics[width=0.98\textwidth]{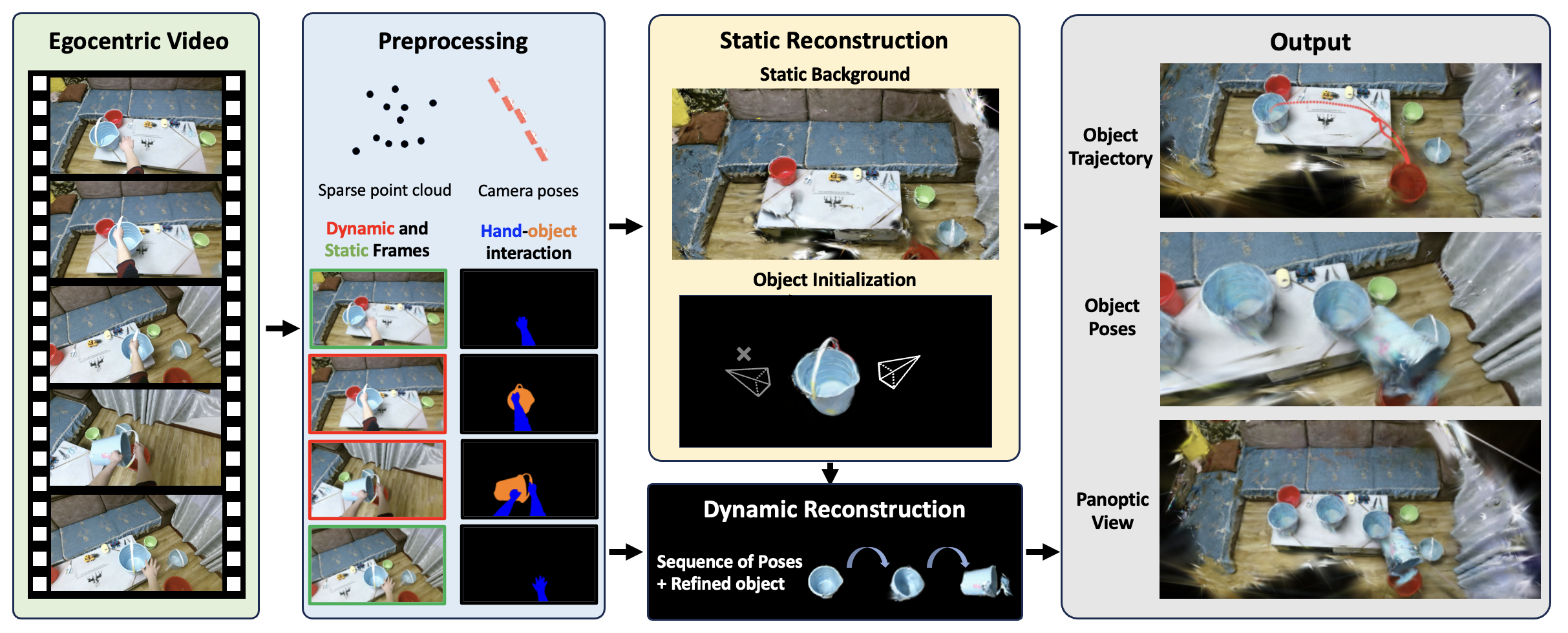}
    \captionof{figure}{\textbf{\methodname Pipeline.} Given an egocentric video input, our framework first partitions the video input into static and dynamic clips and obtains hand-object segmentation masks using off-the-shelf approaches. The static clips are then used to reconstruct the background scenes and initialize the shapes of the object that will be interacted with using 3D Gaussians. Subsequently, we refine the object’s shapes and track their motion through the dynamic clips. \methodname enables a high-quality 4D reconstruction of dynamic scenes with explicit representations that accurately capture the object interactions.} 
    \label{fig:static}
\end{center}%
}]

\begin{abstract}

Human activities are inherently complex, often involving numerous object interactions. 
To better understand these activities, it is crucial to model their interactions with the environment captured through dynamic changes. 
The recent availability of affordable head-mounted cameras and egocentric data offers a more accessible and efficient means to understand human-object interactions in 3D environments. 
However, most existing methods for human activity modeling neglect the dynamic interactions with objects, resulting in only static representations.
The few existing solutions often require inputs from multiple sources, including multi-camera setups, depth-sensing cameras, or kinesthetic sensors. %
To this end, we introduce \methodname, the first method capable of simultaneously reconstructing 3D scenes and dynamically tracking 3D object motion from RGB egocentric input alone.  
We leverage the uniquely discrete nature of Gaussian Splatting and segment dynamic interactions from the background, with both having explicit representations.
Our approach employs a clip-level online learning pipeline that leverages the dynamic nature of human activities, allowing us to reconstruct the temporal evolution of the scene in chronological order and track rigid object motion. 
\methodname shows significant improvements in terms of both dynamic object and background reconstruction quality compared to the state-of-the-art. 
We also qualitatively demonstrate the high quality of the reconstructed models. Video demonstrations, code, and datasets will be available at \href{https://zdwww.github.io/egogs.github.io}{https://zdwww.github.io/egogs.github.io}.
\end{abstract}
    
\section{Introduction}
\label{sec:intro}

Human activities are inherently complex and performing simple household tasks involves numerous interactions with objects. 
For example, making a coffee in the morning involves multiple steps: taking a mug from a shelf, placing it under the coffee machine, pressing a button for the preferred type of coffee, and adding milk or sugar. 
Even this seemingly simple task includes various object interactions and movements. 
To better understand human activities and behaviors, it is important to be able to model these dynamic interactions with the environment. 
The recent availability of affordable head-mounted cameras~\cite{pan2023aria, somasundaram2023project} and egocentric data~\cite{Damen2018EPICKITCHENS, grauman2022ego4d, egoexo4d, lv2024aria} offers a more accessible and efficient means to understand dynamic human-object interactions in 3D environments. 
Toward this goal, we tackle the challenging task of reconstructing 3D scenes and dynamic interactions of objects from RGB egocentric videos. %

Most existing methods for modeling human-object interactions either focus on reconstructing 3D hand-object~\cite{fan2024hold,liu2021semi, wen2023bundlesdf, ye2022hand} or human-scene interaction models~\cite{araujo2023circle, hassan2021populating, Huang_2022_CVPR, lin2016virtual, yi2022mime, zhao2022compositional} or on mapping 3D scenes~\cite{epicfields}. 
These approaches often neglect dynamic interactions with objects, resulting in static representations with motion-induced artifacts, commonly known as the ``ghost effect''. 
The few existing solutions often require inputs from multiple sources, including multi-camera setups \cite{Dynamic3DGaussians}, depth-sensing cameras \cite{wong2021rigidfusion}, or kinesthetic sensors \cite{guzov2022interactionreplica}. 
While these methods achieve 3D reconstruction, they do not consider changes caused by interactions and thus fail to capture the dynamics depicted in egocentric videos.

In this paper, we go beyond prior works to tackle the task of dynamic scene reconstruction from RGB egocentric videos. 
Our proposed method \methodname simultaneously reconstructs 3D scenes and dynamically tracks 3D object motions within them.  
Our key insight is that the uniquely discrete nature of Gaussian Splatting makes it especially suitable for spatial segmentation, allowing objects to be trained separately from the background. 
Given that human activities involve continuous motion over time, we identify critical contact points in time and distinguish dynamic interactions from static captures that only contain camera movements. 
We propose a clip-level online learning pipeline that leverages the dynamic nature of human activities, allowing us to reconstruct the temporal evolution of the scene in chronological order and track rigid object motion.

To reconstruct the dynamic scenes from an egocentric video, \methodname first obtains hand-object segmentation using an off-the-shelf method and derives camera poses through structure-from-motion. 
By leveraging the natural trajectories of interactions, we partition the input video into static and dynamic clips. 
The static clips are used to reconstruct the background scenes and initialize the shapes of the object that will be interacted with. 
Subsequently, we refine the object's shapes and track their motion through the dynamic clips. 
We empirically show that \methodname achieves better reconstruction of dynamic scenes than the state-of-the-art. 
We quantitatively evaluate our method on two in-the-wild egocentric video datasets following the evaluation protocol for novel-view synthesis. 
We also qualitatively demonstrate the high quality of the reconstructed scenes and the tracked object shapes and their motion.

Our main contributions can be summarized as follows: 
\begin{itemize}
    \item %
We present a novel method that accurately reconstructs 3D scenes and dynamic object motion within them from RGB egocentric videos. 
\item We leverage the dynamic nature of interactions that consist of transitions between static and dynamic phases, which facilitates the reconstruction of the static scenes, the object shapes, and the tracking of their motion. 
\item 
Through both qualitative and quantitative evaluation, we demonstrate that our method outperforms previous approaches and provides better 4D reconstruction that captures the dynamic object interactions.  
\end{itemize}

\section{Related Work}
\label{sec:related_work}

\myparagraph{Hand-Object Segmentation}
Many works have studied hand-object interaction in egocentric vision from different aspects. One significant area of focus is segmentation, specifically obtaining image segmentation masks of hands and the objects they hold. Ren et al. \cite{ren2010figure} proposed a motion-based approach to robustly segment both hand and object using optical flow and domain-specific cues from egocentric video. 

Concurrent with the emergence of deep neural networks-based hand-object segmentation is the scaling-up of egocentric data that includes pixel-level annotations and involves diverse daily activities \cite{ Damen2018EPICKITCHENS, Damen2022RESCALING, grauman2022ego4d}. VISOR\cite{darkhalil2022epicvisor} annotates videos from EPIC-KITCHENS\cite{Damen2018EPICKITCHENS, Damen2022RESCALING} dataset and provides masks for 67k hand-object relations covering 36 hours of videos. EgoHOS \cite{zhang2022fineegohos} further introduces the notion of a dense contact boundary to explicitly model the interaction and a context-aware compositional data augmentation technique to generate semantically consistent hand-object segmentation on out-of-distribution egocentric videos. Cheng et al.~\cite{cheng2023towards} produces a rich, unified 2D output of interaction by converting predicted bounding boxes to segments with Segment Anything (SAM) \cite{kirillov2023segmentSAM}. Our method takes egocentric videos with hand-object segmentation masks as input and creates dynamic 3D models. 

\myparagraph{Hand-Object Reconstruction}
Another highly related direction is to reconstruct the hand-object interaction, featuring 3D pose estimation for hands and objects. Recent works often jointly reconstruct hands and objects to favor physically plausible interactions~\cite{ fan2024hold, liu2021semi, wen2023bundlesdf, ye2022hand, ye2023vhoi, zhu2023get}. These approaches can be grouped into two categories. One assumes a known 3D object model and fits that model into 2D image~\cite{cao2021reconstructingRHO, corona2020ganhand, liu2021semi, tekin2019h+, yang2021cpf}. For example, RHO \cite{cao2021reconstructingRHO} adapts an optimization-based approach that's able to reconstruct hands and objects from single images in the wild, by leveraging 2D image cues and 3D contact priors to provide constraints. 

Recent works eliminate the need for a known 3D model and directly reconstruct 3d object shapes from the input~\cite{fan2024hold, qu2023novel, ye2023vhoi}. However, they either require multiview input~\cite{qu2023novel}, specific hand-object interaction supervision~\cite{ye2023vhoi}, or can only reconstruct simple object shapes~\cite{fan2024hold}. Current shape-agnostic methods struggle in in-the-wild scenarios~\cite{fan2024hold, zhu2023get}.
In contrast, our method does not require prior knowledge and obtains 3D object shapes through differentiable 3D Gaussian-based rendering.

\myparagraph{Static Scene Modeling}
In the past few years, the domain of static scene modeling has garnered considerable attention. Mildenhall et al. \cite{original_nerf} introduce the groundbreaking Neural Radiance Fields (NeRF), which utilizes a large Multilayer Perceptron (MLP) to represent 3D scenes and renders via volume rendering technique. However, their method queries the MLP at hundreds of points for each ray, resulting in slow training and rendering speed. Additionally, the original NeRF's performance can diminish in scenes with highly dynamic elements due to its static, volumetric nature. Therefore, some subsequent works have tried to enhance the quality by (1) mitigating existing problems, such as aliasing \cite{mip-nerf, barron2022mipnerf360, mobile-nerf} and reflection \cite{NeRFReN, Ref-NeRF} (2) incorporating image processing \cite{raw-nerf, deblur-nerf} (3) employing per-image transient latent codes \cite{nerf-in-the-wild, GRAF}, and (4) introducing supervision of expected depth with point clouds \cite{DS-NeRF, Point-NeRF}. There also exist some other follow-up works aiming to improve the speed, for example, by caching precomputed MLP results \cite{baked_nerf_1, PlenOctrees}, employing well-designed data structures \cite{TensoRF, DVGO}, removing the neural network \cite{Plenoxels}, or utilizing multi-resolution hash encoding \cite{instant-ngp}. Yet, most of these methods still use ray marching, which involves sampling millions of points and slows down real-time rendering. 

Recently, Kerbl et al. \cite{original_3dgs} propose a different approach in the modeling and rendering of complex static 3D scenes - 3D Gaussian Splatting (3DGS). They model static scenes with Gaussians whose position, opacity, shape, and color are learned through a differentiable splatting-based renderer, achieving real-time rendering speed.

\myparagraph{Dynamic Scene Modeling}
Motivated by the success of NeRF \cite{original_nerf} in static scene modeling, numerous studies have adopted neural representations to model dynamic scenes. One strategy to extend 3D into 4D scenes is by using time stamps as an additional conditioning factor. \cite{dynamic_nerf_per_timestamp_1, dynamic_nerf_per_timestamp_2}. Another set of 'dynamic NeRF' works \cite{HexPlane, TiNeuVox, kplanes, tensor4d, MixVoxels} involves employing 4D space-time grid-based representations. Representing the 3D scene at a certain timestamp as a canonical space and then explicitly modeling deformation fields to warp 3D points into the canonical space is a common strategy as well \cite{df_nerf_4, RDRF, Nerfies, D-NeRF}. Another strategy is to combine the two approaches, using a conditional neural volume together with a deformation field \cite{NeRFSemble, HyperNeRF}.
However, these methods all suffer from the same issues as static NeRFs, in that they require raymarching and despite advances in performance, still are not sufficiently fast for real-time rendering.

Similarly, many dynamic extensions to 3D Gaussian splatting were also proposed \cite{3DDeformableGaussian,  Dynamic3DGaussians, 4DGaussianSplatting, Deformable3DGaussians}. The most common approach is to learn for every timestep a set of deformations for each Gaussians. This can be done explicitly, or implicitly using a deformation which is evaluated for each Gaussian. This results in substantially faster training and rendering speed, with comparable levels of rendering quality.
Although these methods result in decent quality renders, upon closer inspection all of them result in noticeably blurrier results than are possible with static reconstructions, especially when strong motion is involved.

\section{Methodology}

\autoref{fig:static} summarizes our method, \methodname, for dynamic scene reconstruction from RGB egocentric videos. 
\update{Our central idea is to identify dynamic objects and train them separately from the background. 
To do so, we develop a framework that fully integrates the dynamic characteristics of object interactions by temporally segmenting the video into static and dynamic clips and applying existing hand-object interaction modeling techniques to achieve 2D spatial distinction.} 
Specifically, \methodname first obtains camera poses and hand-object segmentation masks and the segmentation masks are further used to partition the videos into static and dynamic clips (Sec.~\ref{sec:preprocessing}).
The static clips are used to initialize the static background and object shapes (Sec.~\ref{sec:staticclip}), while the dynamic clips are used to track object motion and gradually refine their shapes (%
Sec.~\ref{sec:dynamicclip}).

\subsection{Preliminary: 3D Gaussian Splatting}

We use 3D Gaussian Splatting (3D-GS) as our modeling structure because it provides an explicit 3D scene representation with a set of point-cloud-like 3D Gaussians. Each Gaussian is characterized by a position (mean) $\mathbf{\mu}$, a covariance matrix $\Sigma$, an opacity, and color features $\mathbf{c}$. The Gaussians are defined using the standard multivariate Gaussian distribution $G(\mathbf{x})=e^{-\frac{1}{2}(\mathbf{x} - \mathbf{\mu})^T \Sigma^{-1}(\mathbf{x} - \mathbf{\mu})}$, providing a flexible optimization framework and a compact 3D scene representation.

3D Gaussian Splatting utilizes a differentiable point-based $\alpha$-blending rendering to compute the color $C$ of pixel $\mathbf{x}_p$. Specifically, it adapts a typical neural point-based approach and 
blends $N$ ordered points overlapping the pixel:
    $C(\mathbf{x}_p)=\sum_{i \in \mathcal{N}} \mathbf{c}_i \alpha_i \prod_{j=1}^{i-1}\left(1-\alpha_j\right)$, 
where $\alpha_i$ is calculated by evaluating a 2D Gaussian with covariance $\Sigma$, projected from the 3D Gaussian, and then multiplied with its opacity; and $\mathbf{c}_i$ is the color of each Gaussian. The original 3D-GS implementation treats color as a directional appearance component represented via spherical harmonics (SH). For simplicity, we disable the view-dependent color by setting the maximum SH degree to $0$.

\subsection{Data Preprocessing}
\label{sec:preprocessing}
As pointed out by previous work~\cite{gu2024egolifter}, 3D-GS tends to overfit to training views and generate excessive floaters when there are scene inconsistencies among 3D views. In order to eliminate such inconsistencies, our idea is to identify any objects that move at all in the scene and separate them from the static background. 
In this preprocessing step, we adopt an initialisation method similar to other 3D-GS approaches and process the input to separate dynamic contents from static ones.

\myparagraph{Initialisation for 3D-GS}
\label{camera_pose}
\update{Following existing 3D-GS methods, we first use COLMAP~\cite{schoenberger2016sfm, schoenberger2016mvs} to estimate camera poses. }
COLMAP's SfM also creates a sparse point cloud corresponding to the camera poses estimated, and we use them as an initialization for 3D Gaussian Splatting.
For egocentric dataset where camera poses are available, e.g., EPIC Fields \cite{epicfields} provides estimated camera poses for EPIC-KITCHENS \cite{Damen2018EPICKITCHENS}, we employ them directly. 

\myparagraph{Separation of Dynamic and Static}
\update{To separate dynamic and static information in 2D frames, we use off-the-shelf approaches to obtain}
segmentation masks of hand-object interaction.
Specifically, we use %
EgoHOS \cite{zhang2022fineegohos} to generate hand masks and Track-Anything model \cite{yang2023track} for object masks and human body masks. Furthermore, these masks are dilated by 2 pixels for better robustness.
The onset and offset frames of each hand-object interaction are estimated throughout the video. 
We then partition the egocentric video along the temporal axis into \textit{static} and \textit{dynamic} clips according to the onset and offset of interactions. We define static clips as ones where only the actor's hands or body are moving, but objects are all static, while dynamic clips contain both actor and object motion.

\subsection{Static Reconstruction}
\label{sec:staticclip}

\update{Given a static video clip, our goal is to obtain a 3D reconstruction that distinguishes between the static background and dynamic objects involved in interactions within the dynamic clips. 
We begin by training a static representation of the scene and then identify dynamic objects using information extracted from the adjacent dynamic clips. }

\myparagraph{Initial Static Reconstruction} %
We have a set of $T$ observations/frames from the static clip $S = \{ \mathbf{I}_{t}, \mathbf{M}_{\text{body}, t}, \theta_{t} | $ \ $t = 1, \dots, T \}$ as input, where $\mathbf{I}_t$ is an input RGB egocentric frame, $\mathbf{M}_{\text{body}, t}$ is the binary hand/body segmentation mask where pixel value $=0$ represents body part and pixel value $=1$ is for rest of the frame, and $\theta_t$ is the corresponding camera parameters for frame $t$. We follow a similar optimization pipeline as the original 3DGS~\cite{original_3dgs}, including both pruning and densification but use a masked version of the loss function:
\begin{equation*}
\mathcal{L}=(1-\lambda) \mathcal{L}_1 \left(\mathbf{I}_\text{input}, \mathbf{I}_\text{render} \right)
+\lambda \mathcal{L}_{\text {D-SSIM }}\left(\mathbf{I}_\text{input}, \mathbf{I}_\text{render} \right), 
\end{equation*}
with the gradients zeroed out according to the mask $\mathbf{M}_\text{body}$. Similar to SuGaR \cite{guedon2023sugar}, after around 30K iterations, we append an additional entropy loss on the opacity $\alpha$ of Gaussians, i.e.
\begin{equation*}
    \mathcal{L}_{\text{entropy}_\alpha} = -\alpha \log(\alpha) - (1 - \alpha) \log(1 - \alpha), 
    \label{eq:entropy}
\end{equation*}
as a way to enforce Gaussians to be either fully transparent or completely opaque and train for another 10K iterations while disabling pruning and densification. Instead, we prune the transparent Gaussians once at the end of this phase of training.
This produces a set of 3D Gaussians $\mathcal{G}$ reconstructing the scene captured by this static clip, which includes both the static background and any objects that might move during dynamic portions of the video.

\begin{figure*}[h]
  \centering
   \includegraphics[width=1.0\linewidth]{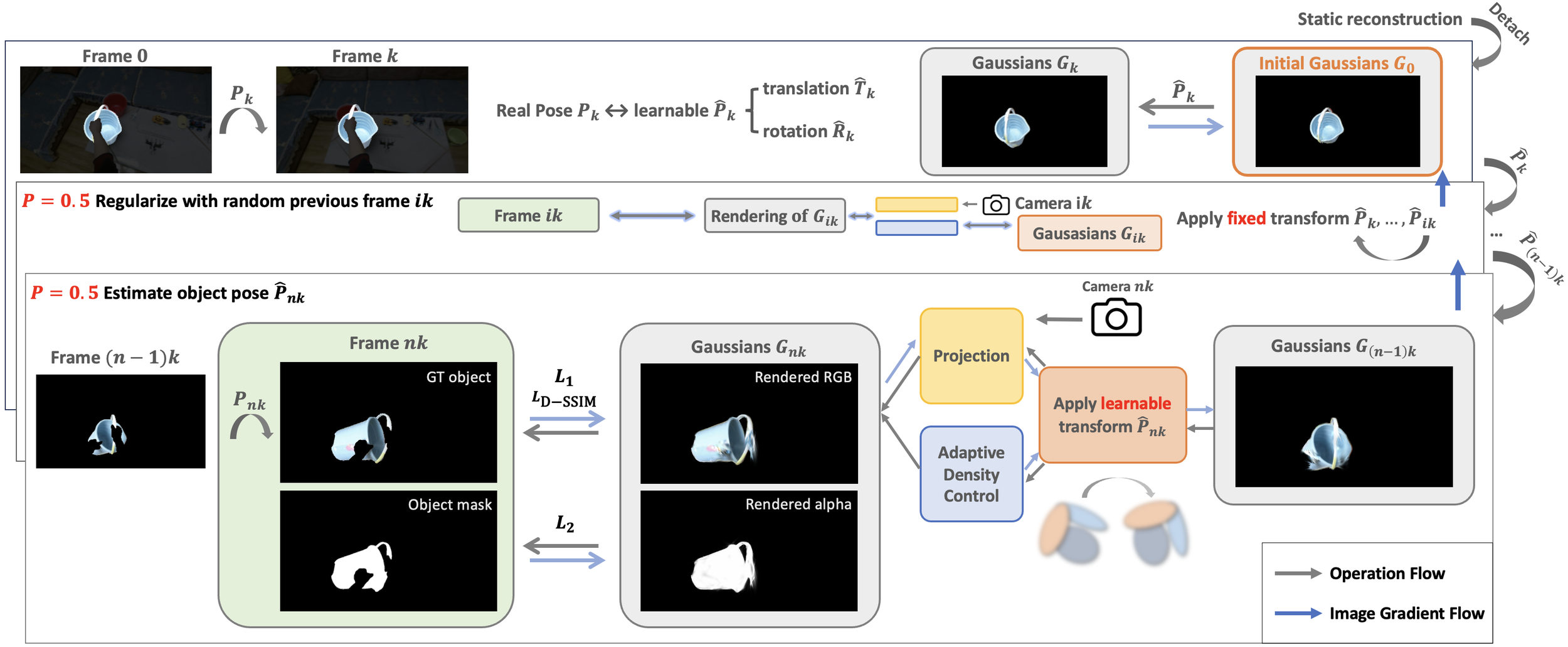}

   \caption{\textbf{Dynamic Object Modeling Pipeline}. We use a sequential pipeline with regularization from previous frames, allowing us to estimate object poses and iteratively refine their shapes simultaneously. 
   }
   \label{fig:dynamic_model}
\end{figure*}

\myparagraph{Dynamic Object Identification} %
\update{To identify the set of Gaussians associated with dynamic objects, we extract interaction information from nearby dynamic clips and aim to generate masks for objects that have moved or will move. 
We automatically generate such masks by selecting a random point from the object mask during interaction within the dynamic clips and using it as a prompt for the Track-Anything model~\cite{yang2023track}.
The initial static reconstruction from the previous step allows us to lift 2D masks to a dense 3D reconstruction of the scene, and conversely, project 3D Gaussian points to 2D. 
Experimental results show that object masks from just $N$ static frames adjacent to the dynamic clips are sufficient to obtain reliable 3D segmentation of the object.} 
For example, consider a static clip $S$ with $T$ frames immediately preceding a dynamic clip where an object is moved by the camera wearer. 
We obtain segmentation masks for the object in the last $N$ frames of this static clip, i.e., $\{ \mathbf{M}_{\text{obj}, T-N}, \dots \mathbf{M}_{\text{obj}, T}\}$, where a pixel value of $=1$ indicates the targeted object and $=0$ represents the rest of the frame. 
We set $N=5$ in our experiments. 

Similar to Gaussian Grouping \cite{ye2023gaussiangrouping}, though more streamline, an additional trainable parameter of label $l$ is then attached to each Gaussian and initialized to a very small value. This label can then be rendered similar to the RGB value as:
   $ L(\mathbf{x}_p)=\sum_{i \in \mathcal{N}} l_i \alpha_i \prod_{j=1}^{i-1}\left(1-\alpha_j\right)$.
This produces a segmentation $\mathbf{L}_\text{r}$, upon which we can apply a binary cross entropy loss using the object masks $\mathbf{M}_{\text{obj}}$:%
\begin{equation*}
    \mathcal{L}_{\text{BCE}_l} = - \left[ \mathbf{M}_\text{obj}  \cdot \ln\left(\sigma(\mathbf{L}_\text{r})\right) + (1- \mathbf{M}_\text{obj}) \cdot \ln\left(1 - \sigma(\mathbf{L}_\text{r})\right) \right]
    \label{eq:BCE}
\end{equation*}
where $\sigma(X) = \frac{1}{1 + e^{-X}}$ is the sigmoid activation function. Training the 3D Gaussians with respect to this loss function while freezing all parameters except for $l$ allows us to separate the Gaussians into object $\mathcal{G}_\text{obj}$ and background $\mathcal{G}_\text{bg}$ based on thresholding $l$ and such 3D segmentation can be used to render targeted object masks from more viewpoints. 
\update{More details and graphical illustration are in the Supp. }

\subsection{Dynamic Reconstruction}
\label{sec:dynamicclip}
Given the reconstructed background $\mathcal{G}_\text{bg}$ and the initial object $\mathcal{G}_\text{obj}$ using the static clips, we refine the object shapes, track their motion through the dynamic clips and update the background with new information revealed by the movement of the dynamic objects. 

\myparagraph{Object Pose Estimation}
We use the set of object Gaussians $\mathcal{G}_\text{obj}$ as an initial estimate of the object appearance 
and estimate its pose for every $k$-th frame in the dynamic clip.
Specifically, we estimate for a targeted frame $f_t$ a corresponding relative pose $\mathbf{P}_t$ from a previous state at $f_{t-k}$. We further decompose the pose $\mathbf{P}_t$ into a 3D translation vector $T_t$ and a rotation matrix $\mathbf{R}_t$. Unlike previous dynamic 3D-GS methods \cite{4DGaussianSplatting, Dynamic3DGaussians}, \methodname applies one set of transformation parameters to the whole collection of object Gaussians $\mathcal{G}_\text{obj}$ as a whole, treating it as a single rigid object. 

We optimize %
the rotation $\mathbf{R}_t$ using the 6D continuous rotation representation $\Tilde{\mathbf{R}}_t$ proposed by \cite{Zhou_2019_CVPR}. 
To ensure the transformation is rigid, an estimated $3 \times 3$ rotation matrix must be $\mathbf{R}_t \in SO(3)$. 
Hence for each target frame $f_t$, %
we apply the estimated %
translation and rotation parameters to the 3D center of each Gaussian:
\begin{equation*}
    \mathbf{X}_{\mathcal{G}_{\text{obj}, t}} = \mathbf{X}_{\mathcal{G}_{\text{obj}, t-k}} \cdot g(\Tilde{\mathbf{R}}_t) + \mathbf{t}_t,
\end{equation*}
where $\mathbf{X}_{\mathcal{G}_{\text{obj}, t}}$ is the 3D coordinates of object Gaussians $\mathcal{G}_{\text{obj}, t}$ %
at time $t$ and $\mathbf{X}_{\mathcal{G}_{\text{obj}, t-k}}$ at time $t-k$. 
$g(\cdot)$ is a function defined in \cite{Zhou_2019_CVPR} that transforms the 6D representation of rotation to a standard $3 \times 3$ rotation matrix. Such object rotation is also applied to the anisotropic covariance of each 3D Gaussian to regularize its alignment with the object surface:
\begin{equation*}
    \Sigma' = \mathbf{R} \Sigma \mathbf{R}^T
    = (\mathbf{R}_t \mathcal{R}) \mathcal{S} \mathcal{S}^T (\mathbf{R}_t \mathcal{R})^T,
\end{equation*}
where the covariance matrix $\Sigma$ is decomposed into the scaling vector $\mathcal{S}$ and the Gaussian's rotation matrix $\mathcal{R}$ constructed from quaternion.

\myparagraph{Object Shape Refinement}
In order to better reconstruct the shape of the object, %
we apply a silhouette loss onto the computed alpha value using the following equation:
$ A(\mathbf{x}_p)=\sum_{i \in \mathcal{N}} \alpha_i \prod_{j=1}^{i-1}\left(1-\alpha_j\right)$,
which effectively equates the RGB rendering equation without color. 

\myparagraph{Dynamic Object Reconstruction}
Our final dynamic object reconstruction loss is then:
\begin{equation*}
    \mathcal{L}_\text{obj}=\mathcal{L}_1 \left(\mathbf{I}_\text{obj}, \mathbf{I}_{\text{render}, \mathcal{G}_\text{obj}}\right)
    + \lambda \mathcal{L}_2 \left( \mathbf{M}_\text{obj}, \mathbf{A}_{\text{render}, \mathcal{G}_\text{obj}} \right), 
\end{equation*}
where $\mathbf{I}_\text{obj}$ is cropped from $\mathbf{I}_\text{input}$ with the object segmentation mask $\mathbf{M}_\text{obj}$. 
$\mathbf{I}_{\text{render}, \mathcal{G}_\text{obj}}$ is rendered from $\mathcal{G}_\text{obj}$ so it contains the object only and black background. 
$\mathbf{A}_{\text{render}, \mathcal{G}_\text{obj}}$ is the rendered alpha. We experimentally observe that $0.5$ is a suitable $\lambda$ value.
Additionally, we lower the learning rate on the Gaussians parameters, such as position and color, by a factor of 10 to prioritize the learning of object poses.

As we optimize the time-dependent pose parameters $T_t$ and $\Tilde{\mathbf{R}}_t$ one frame at a time, the Gaussians can easily overfit to the current frame. To address this, we train not only on the current frame, instead for every training iteration we train either on the current frame or a random previous frame with a probability of $0.5$. 

\myparagraph{Static Scene Update}
The static scene $\mathcal{G}_\text{bg}$ is reconstructed from the static clips, where parts of the background are obscured by objects that interact within the dynamic clips. To utilize the visible information from the dynamic clips, we retrain $\mathcal{G}_\text{bg}$ with the dynamic objects masked out.

\myparagraph{Combination of Static Scene and Dynamic Objects}
As a final step, we combine the object model $\mathcal{G}_\text{obj}$ with the full background model $\mathcal{G}_\text{bg}$. In practice, we note that at this stage, there are often floaters belonging to the background that obscure parts of the object. To eliminate these, we perform a final fine-tuning stage using all training frames and Gaussians. As we focus here on optimizing how the background and dynamic object interact and fit with each other, we again freeze the estimated per-frame object pose. This produces then the full scene reconstruction, including per-frame data of the object pose.

\myparagraph{Training details}
During the Gaussian splatting optimization process, the opacity is frequently set to zero in order to prune floaters. However, this would produce a very noisy signal for the pose optimization. As such, we instead alternate between optimizing the rigid object pose, and densifying/pruning the Gaussians. %
We first train for 4k iterations on every dynamic frame, optimizing the object poses without pruning or densification. We then freeze the object poses, and train another 4k iterations to better incorporate visual information with the estimated object poses, and finally train another 4k iterations optimizing the poses without densification or pruning. For all 12k iterations, all Gaussian parameters such as color and position are continuously optimized.
After iterating through the whole dynamic clip with $M$ frames, we obtain a coarse object pose for each frame $P = \{ T_t, \mathbf{R}_t | t = 1, \hdots, M\}$.
Finally, we perform one final round of joint training using all frames, with 6k iterations of pose estimation, 6k iterations of pruning/densifications, and finally another 6k iterations of pose estimation. This ensures that our object model is more equally fit onto all frames, rather than focused on the last seen ones.

\section{Experiments}

We compare our method with existing baselines for dynamic scene reconstruction where the goal is to reconstruct both static 3D scenes and dynamic objects from RGB egocentric videos. 
To quantitatively assess the quality of the reconstructed 4D scene, we follow the evaluation protocol of the novel view synthesis task using two different egocentric video datasets.
We then present qualitative results of the reconstructed dynamics and conduct ablation studies on two key aspects of the proposed method.

\subsection{Novel View Synthesis}
\label{sec:nvs}

\myparagraph{Datasets}
We evaluate our method on two commonly used egocentric video datasets HOI4D and EPIC-KITCHENS. 

\textit{HOI4D} \cite{HOI4D} is a large-scale egocentric video dataset of human-object interactions consisting of $20$ second-long videos. From this dataset, we randomly select $4$ videos involving rigidly moving objects. Among these, $2$ videos contain mostly translations, while the other $2$ include both translations and rotations. Compared to the original dataset, we downsample the image resolution to one-quarter of its original size, resulting in a resolution of $480$ x $270$ pixels. Each video has a framerate of $15$ FPS.

\textit{EPIC-KITCHENS} \cite{Damen2018EPICKITCHENS} is a large-scale dataset featuring in-the-wild egocentric videos of human-object interactions in native kitchen environments. We again randomly select $4$ video clips involving rigidly moving objects. Of these clips, 2 contain mostly translations, while the other 2 include both translations and rotations. Similar to the HOI4D dataset, we also downsample the image resolution to $455$ x $256$. The average length of these clips is $10.43$ seconds with $60$ FPS.

\myparagraph{Evaluation protocol}
For each video, we train using every second frame and evaluate on the rest. Although we are still able to correctly track and even reconstruct the moving object with fewer training frames (i.e. frames with larger step size), the rapid motion that comes with egocentric videos means that neither our method nor any baselines are able to properly reconstruct the background. We show examples with different step sizes in the Supp.

\myparagraph{Metrics}
To assess the performance of our model, we use the peak signal-to-noise ratio (PSNR), the structural similarity index (SSIM)\cite{1284395}, and the VGG-based perceptual similarity metric LPIPS \cite{zhang2018perceptual}. As we aim to reconstruct the background and object without the actor, we mask out the arm and body of any actors within the scene when computing these metrics and only evaluate the quality of the object and background reconstruction.

\myparagraph{Baselines} 
We compare our method with \update{two state-of-the-art dynamic 3D-GS-based methods, Deformable 3DGS \cite{Deformable3DGaussians} and 4DGS \cite{4DGaussianSplatting}, where both methods apply deformation fields to model monocular dynamic scenes.} 
All compared methods have publicly available codebases, and as such we run the code as provided by the authors on the same preprocessed egocentric data used by our method.

However, unlike our method, both of the SOTA approaches are unable to properly utilize masks of image regions which should not be modeled, in this case the hands of the actor. To ensure a fairer comparison, we additionally modify the compared methods to support masking out gradients on the segmented human body, similar to our method.

\begin{figure*}[h]
\centering
\setkeys{Gin}{width=0.175\textwidth} %

\begin{minipage}[b]{0.02\linewidth}
    \rotatebox{90}{}
\end{minipage}%
\begin{minipage}[b]{0.163\linewidth}
    \centering \small GT
\end{minipage}%
\begin{minipage}[b]{0.163\linewidth}
    \centering \small Ours
\end{minipage}%
\begin{minipage}[b]{0.163\linewidth}
    \centering \small 4DGS \cite{4DGaussianSplatting}
\end{minipage}%
\begin{minipage}[b]{0.163\linewidth}
    \centering \small 4DGS w/o hands
\end{minipage}%
\begin{minipage}[b]{0.163\linewidth}
    \centering \small Def-3DGS \cite{Deformable3DGaussians}
\end{minipage}%
\begin{minipage}[b]{0.163\linewidth}
    \centering \small Def-3DGS w/o hands
\end{minipage}%

\smallskip

\begin{minipage}[b]{0.02\linewidth}
    \rotatebox{90}{\scriptsize HOI Scene1}
\end{minipage}%
\begin{minipage}[b]{0.163\linewidth}
    \includegraphics[width=\linewidth]{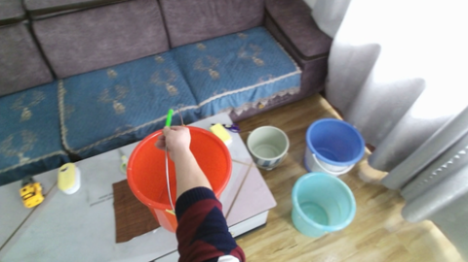}
\end{minipage}%
\begin{minipage}[b]{0.163\linewidth}
    \includegraphics[width=\linewidth]{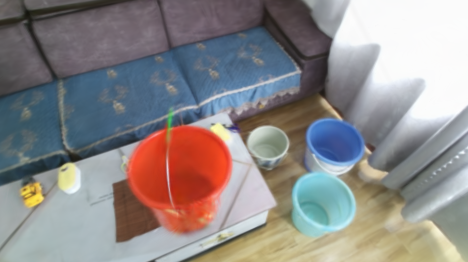}
\end{minipage}%
\begin{minipage}[b]{0.163\linewidth}
    \includegraphics[width=\linewidth]{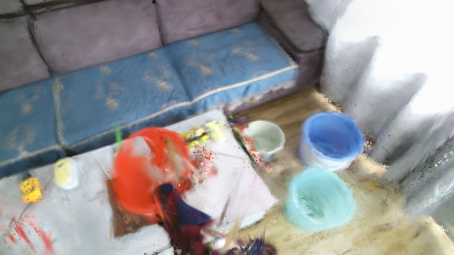}
\end{minipage}%
\begin{minipage}[b]{0.163\linewidth}
    \includegraphics[width=\linewidth]{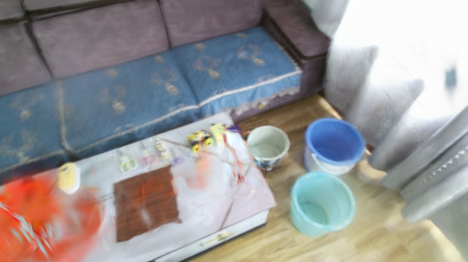}
\end{minipage}%
\begin{minipage}[b]{0.163\linewidth}
    \includegraphics[width=\linewidth]{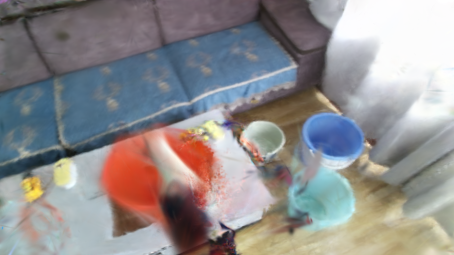}
\end{minipage}%
\begin{minipage}[b]{0.163\linewidth}
    \includegraphics[width=\linewidth]{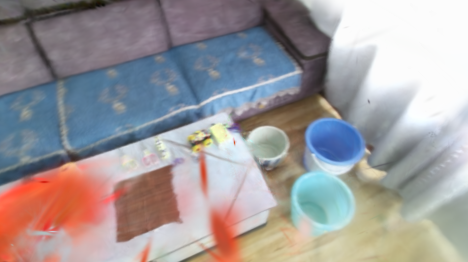}
\end{minipage}%

\begin{minipage}[b]{0.02\linewidth}
    \rotatebox{90}{\scriptsize HOI Scene2}
\end{minipage}%
\begin{minipage}[b]{0.163\linewidth}
    \includegraphics[width=\linewidth]{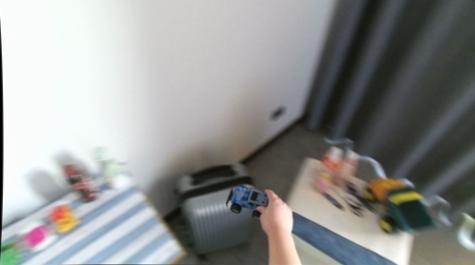}
\end{minipage}%
\begin{minipage}[b]{0.163\linewidth}
    \includegraphics[width=\linewidth]{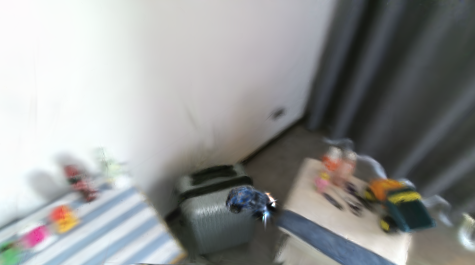}
\end{minipage}%
\begin{minipage}[b]{0.163\linewidth}
    \includegraphics[width=\linewidth]{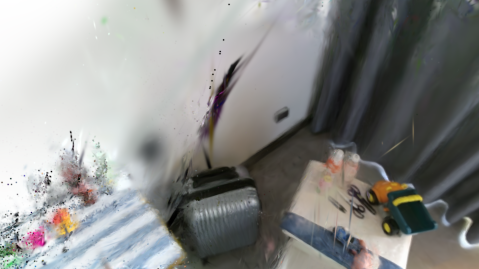}
\end{minipage}%
\begin{minipage}[b]{0.163\linewidth}
    \includegraphics[width=\linewidth]{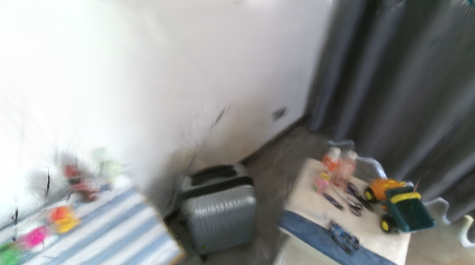}
\end{minipage}%
\begin{minipage}[b]{0.163\linewidth}
    \includegraphics[width=\linewidth]{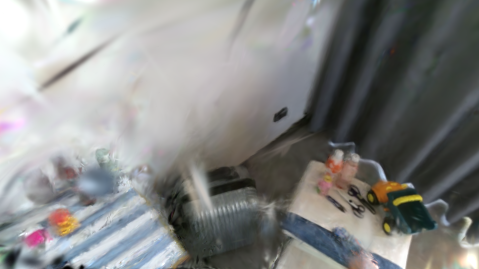}
\end{minipage}%
\begin{minipage}[b]{0.163\linewidth}
    \includegraphics[width=\linewidth]{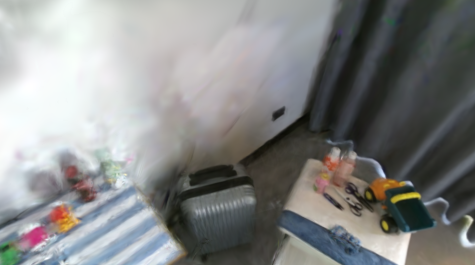}
\end{minipage}%

\begin{minipage}[b]{0.02\linewidth}
    \rotatebox{90}{\scriptsize HOI Scene3}
\end{minipage}%
\begin{minipage}[b]{0.163\linewidth}
    \includegraphics[width=\linewidth]{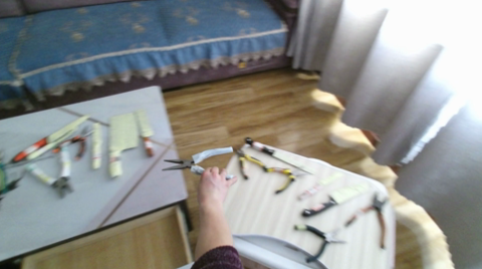}
\end{minipage}%
\begin{minipage}[b]{0.163\linewidth}
    \includegraphics[width=\linewidth]{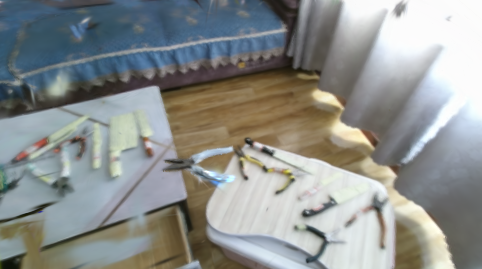}
\end{minipage}%
\begin{minipage}[b]{0.163\linewidth}
    \includegraphics[width=\linewidth]{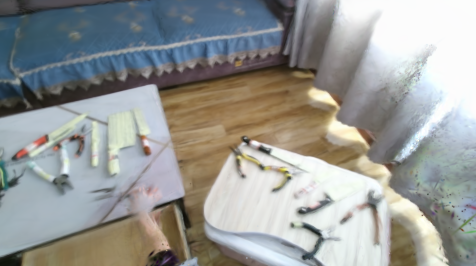}
\end{minipage}%
\begin{minipage}[b]{0.163\linewidth}
    \includegraphics[width=\linewidth]{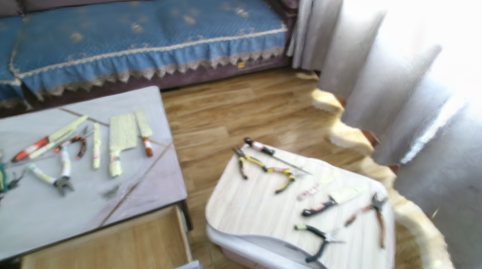}
\end{minipage}%
\begin{minipage}[b]{0.163\linewidth}
    \includegraphics[width=\linewidth]{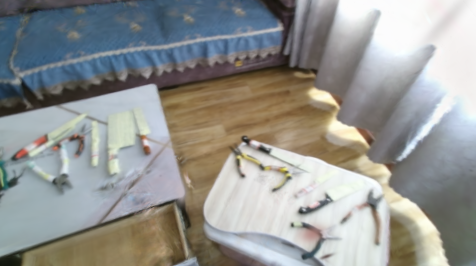}
\end{minipage}%
\begin{minipage}[b]{0.163\linewidth}
    \includegraphics[width=\linewidth]{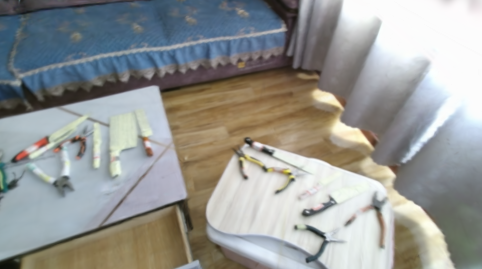}
\end{minipage}%

\begin{minipage}[b]{0.02\linewidth}
    \rotatebox{90}{\scriptsize HOI Scene4}
\end{minipage}%
\begin{minipage}[b]{0.163\linewidth}
    \includegraphics[width=\linewidth]{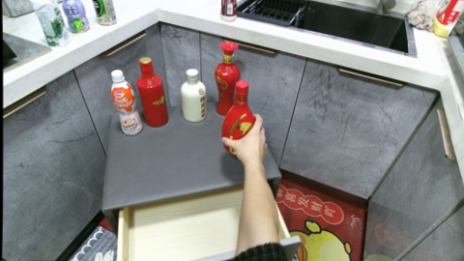}
\end{minipage}%
\begin{minipage}[b]{0.163\linewidth}
    \includegraphics[width=\linewidth]{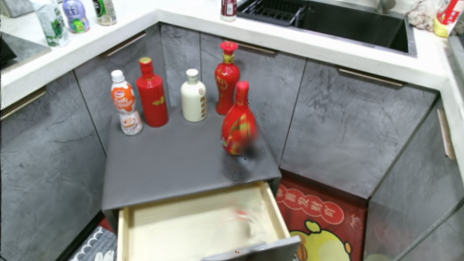}
\end{minipage}%
\begin{minipage}[b]{0.163\linewidth}
    \includegraphics[width=\linewidth]{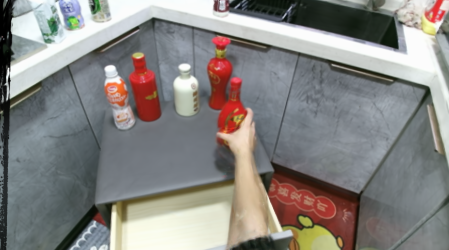}
\end{minipage}%
\begin{minipage}[b]{0.163\linewidth}
    \includegraphics[width=\linewidth]{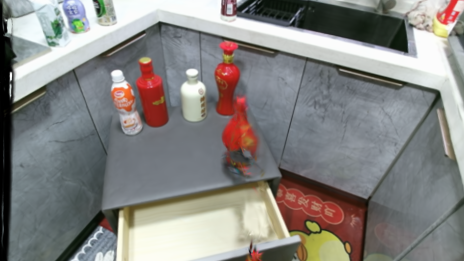}
\end{minipage}%
\begin{minipage}[b]{0.163\linewidth}
    \includegraphics[width=\linewidth]{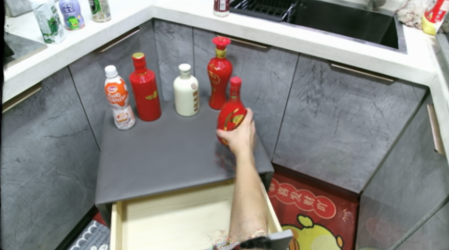}
\end{minipage}%
\begin{minipage}[b]{0.163\linewidth}
    \includegraphics[width=\linewidth]{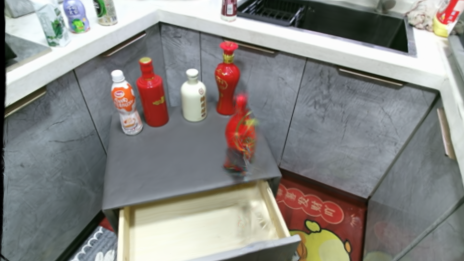}
\end{minipage}%

\begin{minipage}[b]{0.02\linewidth}
    \rotatebox{90}{\scriptsize EK Scene1}
\end{minipage}%
\begin{minipage}[b]{0.163\linewidth}
    \includegraphics[width=\linewidth]{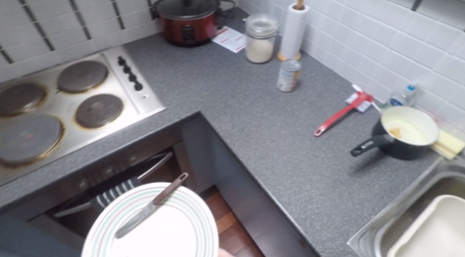}
\end{minipage}%
\begin{minipage}[b]{0.163\linewidth}
    \includegraphics[width=\linewidth]{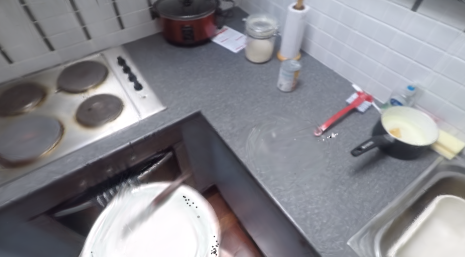}
\end{minipage}%
\begin{minipage}[b]{0.163\linewidth}
    \includegraphics[width=\linewidth]{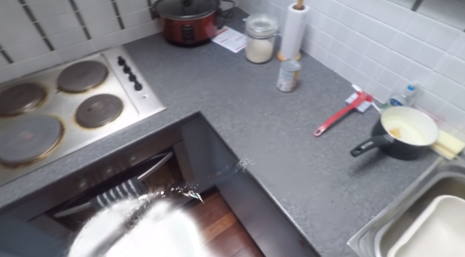}
\end{minipage}%
\begin{minipage}[b]{0.163\linewidth}
    \includegraphics[width=\linewidth]{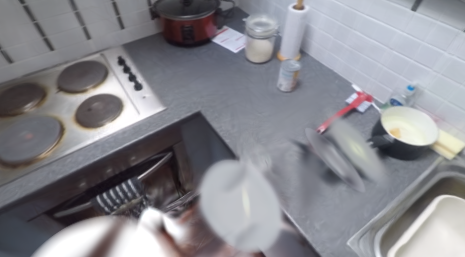}
\end{minipage}%
\begin{minipage}[b]{0.163\linewidth}
    \includegraphics[width=\linewidth]{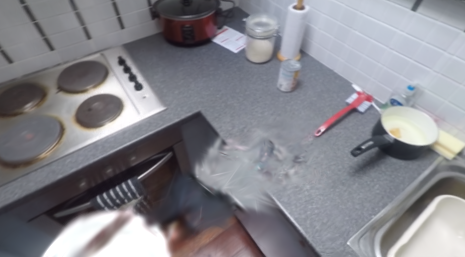}
\end{minipage}%
\begin{minipage}[b]{0.163\linewidth}
    \includegraphics[width=\linewidth]{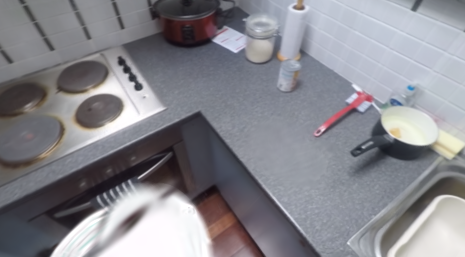}
\end{minipage}%

\begin{minipage}[b]{0.02\linewidth}
    \rotatebox{90}{\scriptsize EK Scene2}
\end{minipage}%
\begin{minipage}[b]{0.163\linewidth}
    \includegraphics[width=\linewidth]{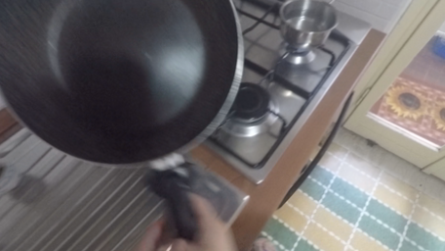}
\end{minipage}%
\begin{minipage}[b]{0.163\linewidth}
    \includegraphics[width=\linewidth]{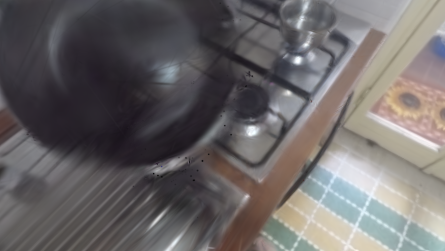}
\end{minipage}%
\begin{minipage}[b]{0.163\linewidth}
    \includegraphics[width=\linewidth]{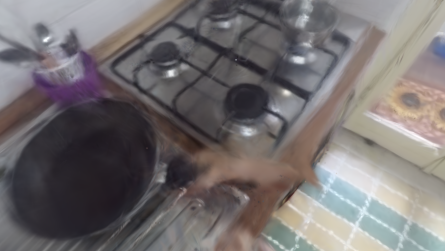}
\end{minipage}%
\begin{minipage}[b]{0.163\linewidth}
    \includegraphics[width=\linewidth]{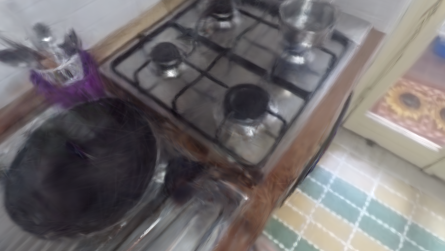}
\end{minipage}%
\begin{minipage}[b]{0.163\linewidth}
    \includegraphics[width=\linewidth]{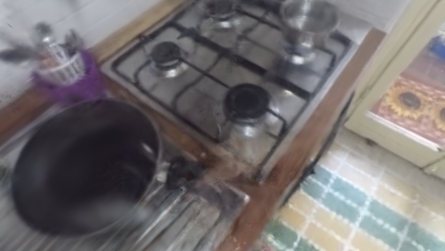}
\end{minipage}%
\begin{minipage}[b]{0.163\linewidth}
    \includegraphics[width=\linewidth]{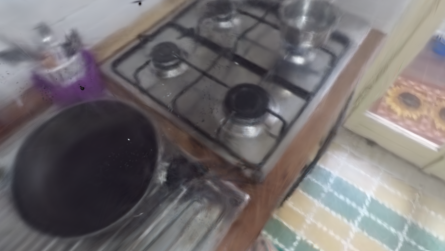}
\end{minipage}%

\caption{\textbf{Qualitative comparison with SOTA.} We show reconstructions produced by our method and SOTA baselines (4DGS \cite{4DGaussianSplatting} and Deformable 3DGS \cite{Deformable3DGaussians}) along with their modified versions from both HOI4D and EPIC-KITCHENS. Our reconstruction demonstrates more accurate reconstructions while baseline approaches fail to handle dynamic interactions.}
\label{fig:exp_41}
\vspace{-0.5cm}
\end{figure*}

\myparagraph{Results}
\autoref{tab:nvs} and \autoref{fig:exp_41} compares our method with existing dynamic 3D-GS methods and their modified versions. We observe that, though the adaptation of gradient masks increases the performance of existing methods, \methodname still significantly outperforms both methods on all evaluation metrics over the dynamic frames of two datasets, with the two SOTA methods performing similarly on both datasets. 

The supplementary video shows the sequences from which the images originate separately. The video results suggest that the deformation-field-based method is not able to track the rigid movement of objects in egocentric video and tends to overfit to the training views. 

\begin{table*}[t]
\centering
\setlength\tabcolsep{2pt} %
\resizebox{0.95\linewidth}{!}{%
\begin{tabular}{@{}cccccccccccccccc@{}}
\toprule
\multicolumn{1}{l}{} & \multicolumn{7}{c}{\textbf{HOI4D}} & \multicolumn{1}{l}{} & \multicolumn{7}{c}{\textbf{Epic-Kitchen}} \\
\textbf{Method} &
  \multicolumn{3}{c}{\textbf{Static}} &
  \multicolumn{1}{l}{} &
  \multicolumn{3}{c}{\textbf{Dynamic}} &
  \multicolumn{1}{l}{} &
  \multicolumn{3}{c}{\textbf{Static}} &
  \multicolumn{1}{l}{} &
  \multicolumn{3}{c}{\textbf{Dynamic}} \\ \cmidrule(lr){2-4} \cmidrule(lr){6-8} \cmidrule(lr){10-12} \cmidrule(l){14-16} 
\multicolumn{1}{l}{} &
  \multicolumn{1}{l}{SSIM $\uparrow$} &
  \multicolumn{1}{l}{PSNR $\uparrow$} &
  \multicolumn{1}{l}{LPIPS $\downarrow$} &
  \multicolumn{1}{l}{} &
  \multicolumn{1}{l}{SSIM $\uparrow$} &
  \multicolumn{1}{l}{PSNR $\uparrow$} &
  \multicolumn{1}{l}{LPIPS $\downarrow$} &
  \multicolumn{1}{l}{} &
  \multicolumn{1}{l}{SSIM $\uparrow$} &
  \multicolumn{1}{l}{PSNR $\uparrow$} &
  \multicolumn{1}{l}{LPIPS $\downarrow$} &
  \multicolumn{1}{l}{} &
  \multicolumn{1}{l}{SSIM $\uparrow$} &
  \multicolumn{1}{l}{PSNR $\uparrow$} &
  \multicolumn{1}{l}{LPIPS $\downarrow$} \\ \midrule

4DGS~\cite{4DGaussianSplatting}           & 0.88 & 25.33 & 0.13 & & 0.89 & 25.34 & 0.13 & & 0.84 & 26.84 & 0.20 & & 0.79 & 22.54 & 0.24   \\
4DGS w/o hands & \textit{0.94} & \textit{28.69} & \textit{0.08} & & \textit{0.94} & \textit{27.33} & \textit{0.10} & & \textbf{0.87} & \textbf{28.90} & \textbf{0.16} & & 0.80 & 23.13 & 0.23 \\
Def-3DGS~\cite{Deformable3DGaussians}     & 0.90 & \textit{25.85} & \textit{0.11} & & \textit{0.90} & 25.71 & 0.12 & & \textit{0.86} & \textit{27.35} & 0.18 & & 0.81 & \textit{23.15} & 0.22    \\
Def-3DGS w/o hands & \textit{0.94} & 28.09 & \textit{0.08} & & \textit{0.94} & 26.92 & \textit{0.10} & & \textit{0.86} & 27.63 & \textit{0.17} & & \textit{0.82} & \textit{23.27} & \textit{0.21} \\
Ours                 & \textbf{0.96} & \textbf{30.99} & \textbf{0.08} & & \textbf{0.95} & \textbf{30.33} & \textbf{0.09} & & 0.85 & \textit{28.33} & 0.19 & & \textbf{0.88} & \textbf{28.34} & \textbf{0.17}
\\ \bottomrule
\end{tabular}%
}

\caption{\textbf{Comparison with SOTA dynamic Gaussian Splatting methods.} We evaluate our method and two other SOTA baselines along with their modified versions with hands excluded from modeling on the HOI4D and EK datasets. The best and second best results are \textbf{bolded} and \textit{italicized} respectively. We show the evaluation results on static and dynamic frames separately.}
\label{tab:nvs}
\end{table*}

\subsection{Dynamic modeling}

\begin{figure}[h]
\centering
\setkeys{Gin}{width=0.98\textwidth} %

\begin{minipage}[b]{0.58\linewidth}
    \includegraphics[width=\linewidth]{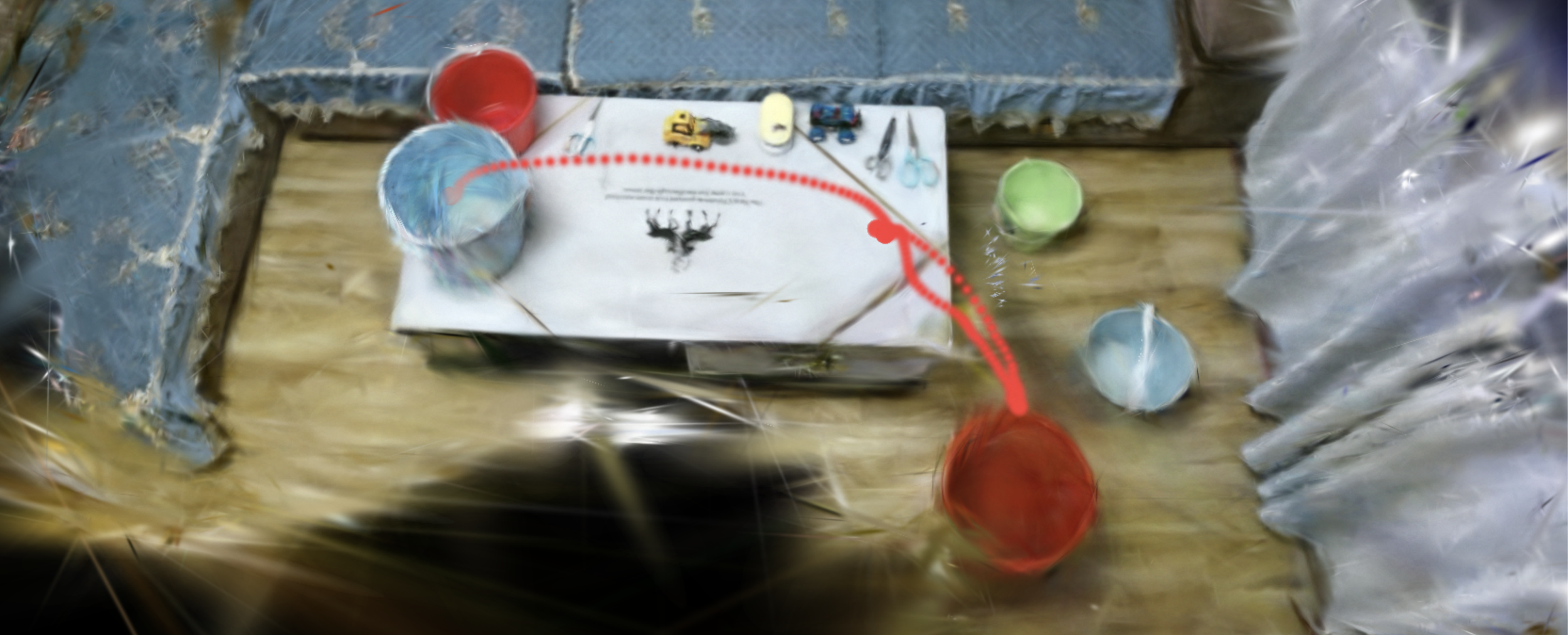}
\end{minipage}%
\begin{minipage}[b]{0.42\linewidth}
    \includegraphics[width=\linewidth]{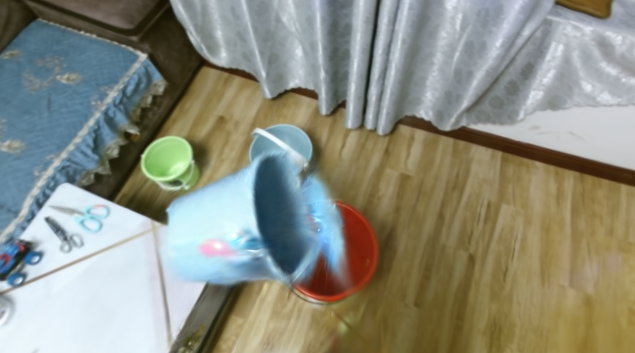}
\end{minipage}%

\begin{minipage}[b]{0.58\linewidth}
    \includegraphics[width=\linewidth]{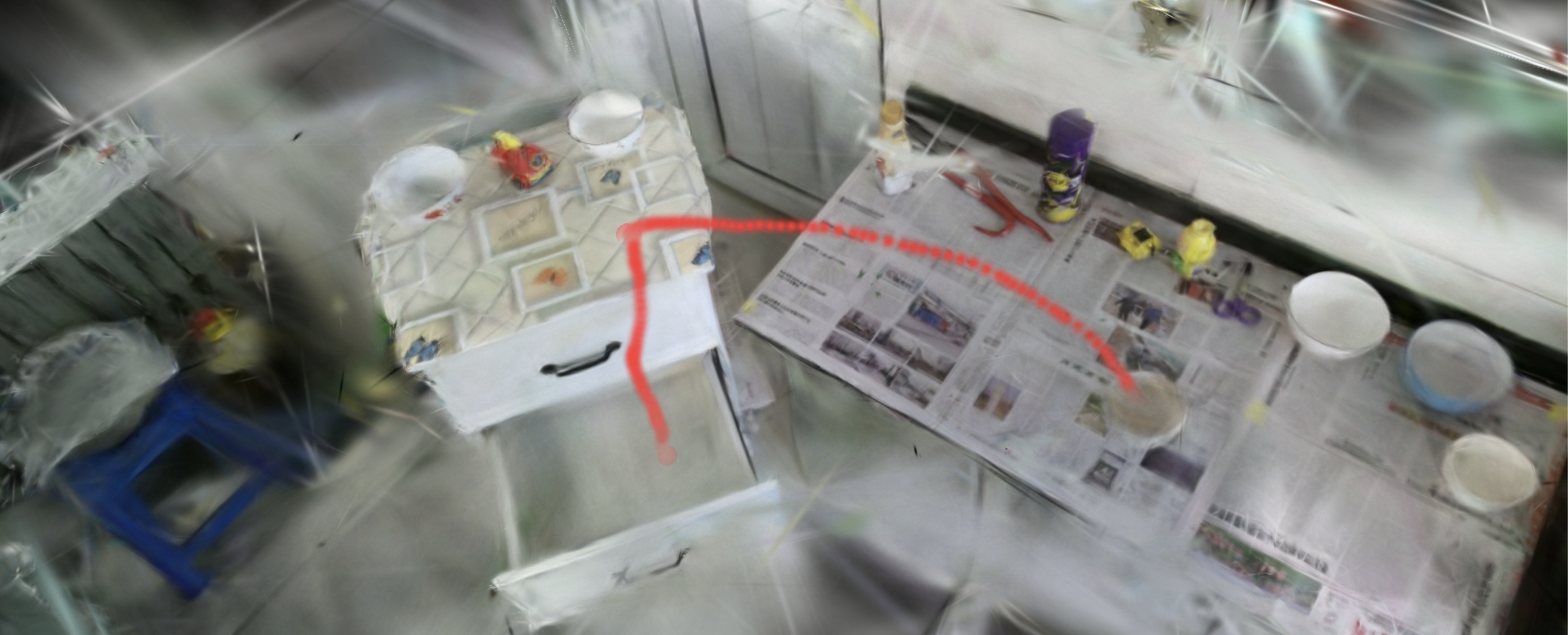}
\end{minipage}%
\begin{minipage}[b]{0.42\linewidth}
    \includegraphics[width=\linewidth]{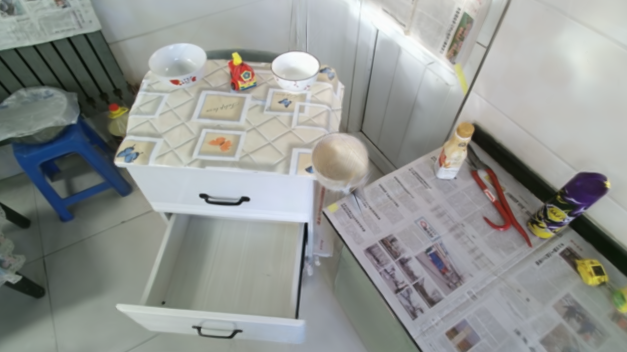}
\end{minipage}%

\begin{minipage}[b]{0.58\linewidth}
    \includegraphics[width=\linewidth]{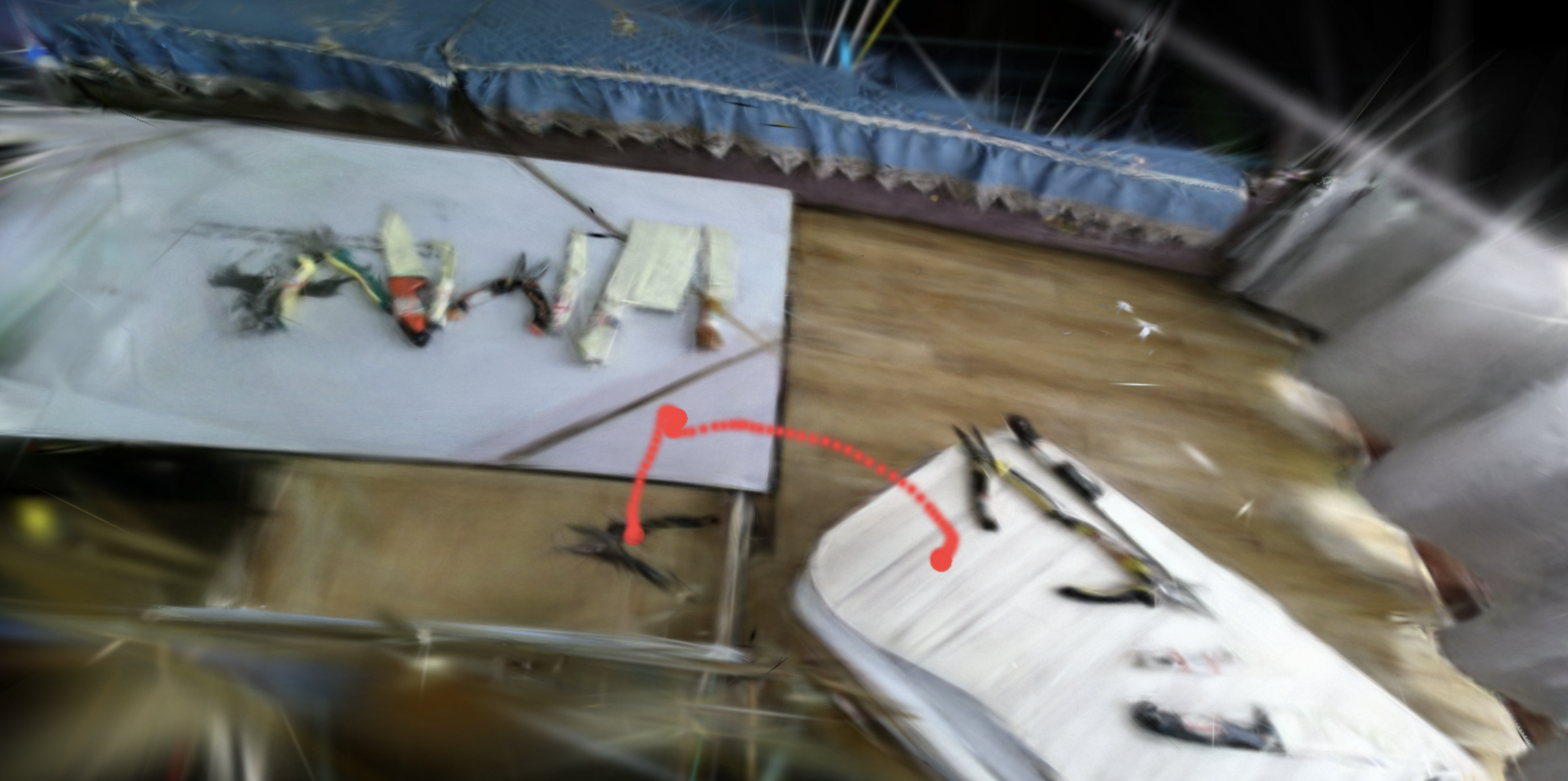}
\end{minipage}%
\begin{minipage}[b]{0.42\linewidth}
    \includegraphics[width=\linewidth]{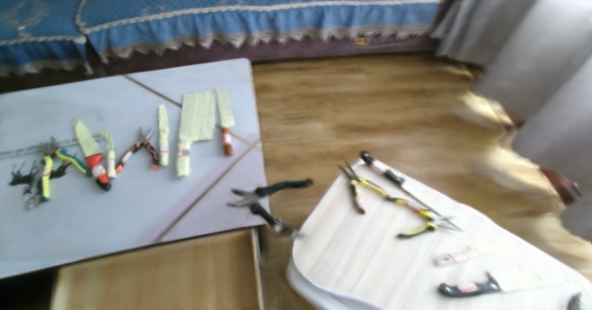}
\end{minipage}%

\begin{minipage}[b]{0.58\linewidth}
    \includegraphics[width=\linewidth]{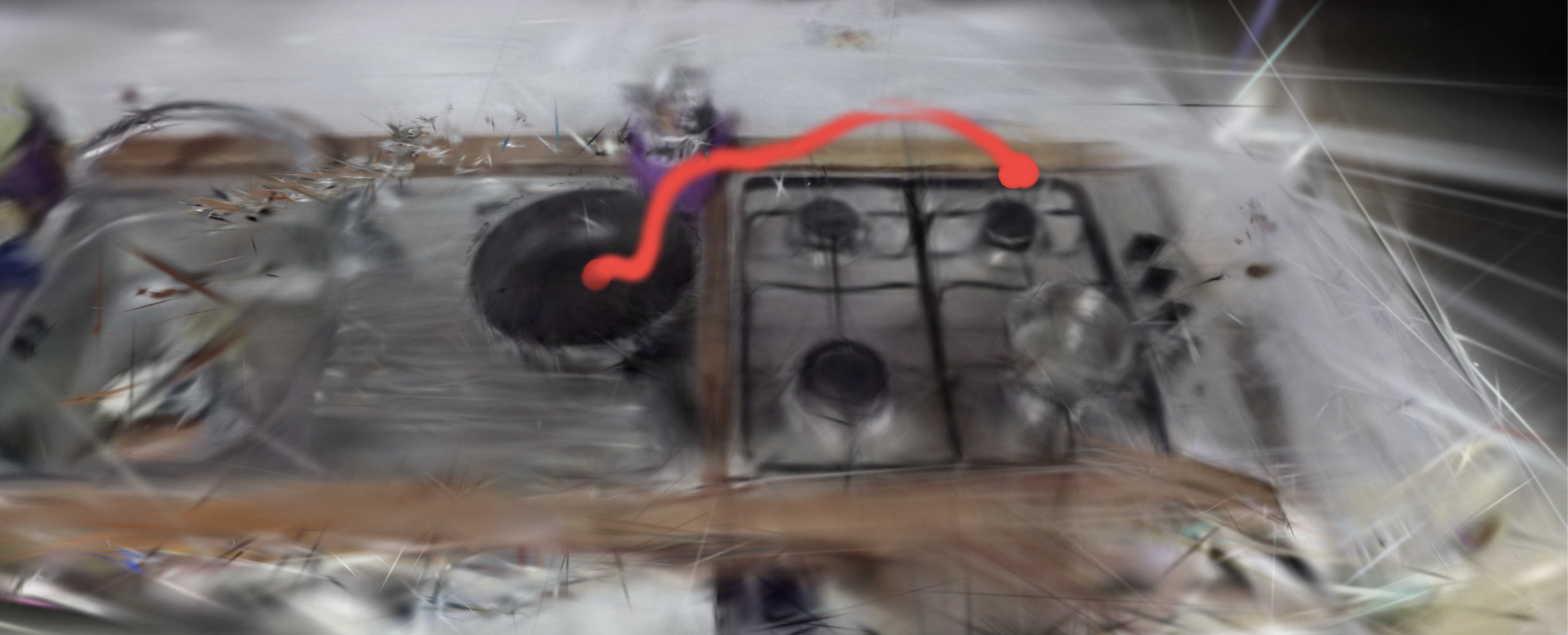}
\end{minipage}%
\begin{minipage}[b]{0.42\linewidth}
    \includegraphics[width=\linewidth]{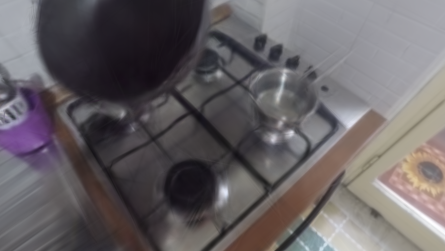}
\end{minipage}

\caption{\textbf{More qualitative results of the reconstructed dynamic scenes.} The left figure shows the object trajectory inferred from interpolated object poses. The right figure shows the rendering from arbitrary viewpoints.
}
\label{fig:qualitative}
\end{figure}

\autoref{fig:qualitative} shows the estimated object trajectories and novel views rendered from arbitrary viewpoints. 
Demonstration videos and more visualizations are included in the Supp.

\begin{table*}[t]
\centering
\setlength\tabcolsep{2pt} %
\resizebox{0.85\linewidth}{!}{%
\begin{tabular}{@{}cccclccclccclccc@{}}
\toprule
\multicolumn{1}{l}{} &
  \multicolumn{7}{c}{\textbf{HOI4D}} &
   &
  \multicolumn{7}{c}{\textbf{Epic-Kitchen}} \\
\textbf{Method} &
  \multicolumn{3}{c}{\textbf{Static}} &
   &
  \multicolumn{3}{c}{\textbf{Dynamic}} &
   &
  \multicolumn{3}{c}{\textbf{Static}} &
   &
  \multicolumn{3}{c}{\textbf{Dynamic}} \\ \cmidrule(lr){2-4} \cmidrule(lr){6-8} \cmidrule(lr){10-12} \cmidrule(l){14-16} 
\multicolumn{1}{l}{} &
  \multicolumn{1}{l}{SSIM $\uparrow$} &
  \multicolumn{1}{l}{PSNR $\uparrow$} &
  \multicolumn{1}{l}{LPIPS $\downarrow$} &
   &
  \multicolumn{1}{l}{SSIM $\uparrow$} &
  \multicolumn{1}{l}{PSNR $\uparrow$} &
  \multicolumn{1}{l}{LPIPS $\downarrow$} &
   &
  \multicolumn{1}{l}{SSIM $\uparrow$} &
  \multicolumn{1}{l}{PSNR $\uparrow$} &
  \multicolumn{1}{l}{LPIPS $\downarrow$} &
   &
  \multicolumn{1}{l}{SSIM $\uparrow$} &
  \multicolumn{1}{l}{PSNR $\uparrow$} &
  \multicolumn{1}{l}{LPIPS $\downarrow$} \\ \midrule
With Original Step Size &
  \textbf{0.98} &
  \textbf{31.82} &
  \textbf{0.03} &
  \multicolumn{1}{c}{} &
  \textbf{0.98} &
  \textbf{29.79} &
  0.04 &
  \multicolumn{1}{c}{} &
  \textbf{0.97} &
  \textbf{28.87} &
  \textbf{0.05} &
  \multicolumn{1}{c}{} &
  \textbf{0.97} &
  \textbf{31.33} &
  \textbf{0.05} \\
With Larger Step Size &
  \textbf{0.98} &
  28.61 &
  \textbf{0.03} &
  \multicolumn{1}{c}{} &
  \textbf{0.98} &
  28.35 &
  \textbf{0.03} &
  \multicolumn{1}{c}{} &
  0.94 &
  24.01 &
  0.06 &
  \multicolumn{1}{c}{} &
  0.92 &
  26.68 &
  0.08 \\ \bottomrule
\end{tabular}%
}
\caption{\textbf{Ablation study} of step size over the object on object reconstruction.}
\label{tab:ablation_step_size}
\end{table*}

\begin{table*}[t]
\centering
\setlength\tabcolsep{2pt} %
\resizebox{0.85\linewidth}{!}{%
\begin{tabular}{@{}cccclccclccclccc@{}}
\toprule
\multicolumn{1}{l}{} &
  \multicolumn{7}{c}{\textbf{HOI4D}} &
   &
  \multicolumn{7}{c}{\textbf{Epic-Kitchen}} \\
\textbf{Method} &
  \multicolumn{3}{c}{\textbf{Static}} &
   &
  \multicolumn{3}{c}{\textbf{Dynamic}} &
   &
  \multicolumn{3}{c}{\textbf{Static}} &
   &
  \multicolumn{3}{c}{\textbf{Dynamic}} \\ \cmidrule(lr){2-4} \cmidrule(lr){6-8} \cmidrule(lr){10-12} \cmidrule(l){14-16} 
\multicolumn{1}{l}{} &
  \multicolumn{1}{l}{SSIM $\uparrow$} &
  \multicolumn{1}{l}{PSNR $\uparrow$} &
  \multicolumn{1}{l}{LPIPS $\downarrow$} &
   &
  \multicolumn{1}{l}{SSIM $\uparrow$} &
  \multicolumn{1}{l}{PSNR $\uparrow$} &
  \multicolumn{1}{l}{LPIPS $\uparrow$} &
   &
  \multicolumn{1}{l}{SSIM $\uparrow$} &
  \multicolumn{1}{l}{PSNR $\uparrow$} &
  \multicolumn{1}{l}{LPIPS $\downarrow$} &
   &
  \multicolumn{1}{l}{SSIM $\uparrow$} &
  \multicolumn{1}{l}{PSNR $\uparrow$} &
  \multicolumn{1}{l}{LPIPS $\downarrow$} \\ \midrule
With Fine-tuning &
  \textbf{0.96} &
  \textbf{31.52} &
  \textbf{0.07} &
  \multicolumn{1}{c}{} &
  \textbf{0.95} &
  \textbf{30.29} &
  \textbf{0.09} &
  \multicolumn{1}{c}{} &
  \textbf{0.94} &
  \textbf{34.22} &
  \textbf{0.10} &
  \multicolumn{1}{c}{} &
  \textbf{0.88} &
  \textbf{28.30} &
  \textbf{0.17} \\
Without Fine-tuning &
  0.87 &
  23.90 &
  0.15 &
  \multicolumn{1}{c}{} &
  0.86 &
  23.03 &
  0.17 &
  \multicolumn{1}{c}{} &
  0.78 &
  21.58 &
  0.24 &
  \multicolumn{1}{c}{} &
  0.79 &
  21.04 &
  0.24 \\ \bottomrule
\end{tabular}%
}
\caption{\textbf{Ablation study} of full scene fine tuning.}
\label{tab:ablation_fine_tuning}
\end{table*}

\subsection{Ablation study}

\begin{figure}[h]
\centering
\begin{minipage}[b]{0.25\linewidth}
    \includegraphics[width=\linewidth]{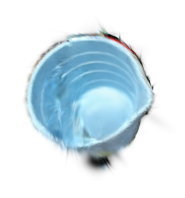}
\end{minipage}%
\begin{minipage}[b]{0.25\linewidth}
    \includegraphics[width=\linewidth]{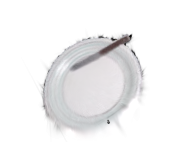}
\end{minipage}%
\begin{minipage}[b]{0.25\linewidth}
    \includegraphics[width=\linewidth]{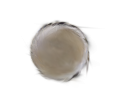}
\end{minipage}%

\caption{Example of black artifacts}
\label{fig:ablation_deform2}
\end{figure}

\myparagraph{Estimate poses with larger time gap}
We show that our chronological pose estimation schema is also able to model the dynamic object with larger time gaps $k$, by training on every 6 frames instead of every 2 frames. We can then estimate the state of object at each timestamp $t$ through interpolation of transformation matrix constructed from relative pose $\mathbf{P}_t$ from $t-k$ to $t$.
As seen in Table \ref{tab:ablation_step_size}, although PSNR drops, we are still nonetheless able to produce an accurate reconstruction. Note that the metrics are computed only on the dynamic object itself.

\myparagraph{Without full scene fine-tuning}
We show the necessity of fine-tuning the static background and dynamic object as described in Section \ref{sec:dynamicclip} jointly by comparing how our method performs when the background and object are only trained in isolation without fine-tuning on all frames or on the combined scene. As can be seen in Table \ref{tab:ablation_fine_tuning} without full scene fine-tuning, quality drops significantly. This is partially due to 3D-GS being unable to distinguish between transparency and blackness. By jointly training both object and background, we eliminate this uncertainty and such black artifacts. We show examples of this in \autoref{fig:ablation_deform2}.

\section{Conclusion and Discussion}

We introduced EgoGaussian, a novel egocentric reconstruction method that is able to reconstruct rigid objects along with accompanying motion from egocentric data. We show significant improvements in terms of both dynamic object and background reconstruction quality compared to the state-of-the-art.

Although our method is able to well reconstruct rapid, rigid object motion, there are still a number of important limitations.
    As we require labels for both onset and offset of objects motion, as well as object masks, we require several additional offline data-preprocessing steps. 
    Our method fundamentally relies on multiview-stereo data in order to reconstruct the scene geometry. As such, our method can encounter overfitting if the datasets have limited viewpoint coverage or lack features that can be tracked across time or view (e.g. uniformly textured walls).
    We utilize pixel-wise gradients to estimate per-frame object motion. As such, if the object goes out of frame, or there is too much motion between two frames, our method can lose track of the object entirely.
    The frame-by-frame dynamic object modelling requires significantly longer training compared to most previous dynamic 3D-GS-based methods.       

Both motion labels and mask segmentations are parallel avenues of ongoing research, and as improvements are made, can easily be integrated into our pipeline.
Similarly, as our method fundamentally trains using static or rigid sets of Gaussians, future improvements in sparse 3D Gaussian reconstruction could also be integrated into our pipeline.
Although pixel-wise gradients prove to be sufficient in our setting, non-local methods of supervision such as optical flow would be an interesting avenue of research in order to improve robustness for large motions or when objects move out of frame. Such forms of supervision could also be utilized in order to estimate per-frame object pose more rapidly, allowing for reduced training time.

\clearpage
{
    \small
    \bibliographystyle{ieeenat_fullname}

\begin{thebibliography}{78}
        \providecommand{\natexlab}[1]{#1}
        \providecommand{\url}[1]{\texttt{#1}}
        \expandafter\ifx\csname urlstyle\endcsname\relax
          \providecommand{\doi}[1]{doi: #1}\else
          \providecommand{\doi}{doi: \begingroup \urlstyle{rm}\Url}\fi
        
        \bibitem[Ara{\'u}jo et~al.(2023)Ara{\'u}jo, Li, Vetrivel, Agarwal, Wu, Gopinath, Clegg, and Liu]{araujo2023circle}
        Joao~Pedro Ara{\'u}jo, Jiaman Li, Karthik Vetrivel, Rishi Agarwal, Jiajun Wu, Deepak Gopinath, Alexander~William Clegg, and Karen Liu.
        \newblock Circle: Capture in rich contextual environments.
        \newblock In \emph{Proceedings of the IEEE/CVF Conference on Computer Vision and Pattern Recognition}, pages 21211--21221, 2023.
        
        \bibitem[Bansal and Zollhoefer(2023)]{dynamic_nerf_per_timestamp_1}
        A. Bansal and M. Zollhoefer.
        \newblock Neural pixel composition for 3d-4d view synthesis from multi-views.
        \newblock In \emph{2023 IEEE/CVF Conference on Computer Vision and Pattern Recognition (CVPR)}, pages 290--299, Los Alamitos, CA, USA, 2023. IEEE Computer Society.
        
        \bibitem[Barron et~al.(2021)Barron, Mildenhall, Tancik, Hedman, Martin-Brualla, and Srinivasan]{mip-nerf}
        Jonathan~T. Barron, Ben Mildenhall, Matthew Tancik, Peter Hedman, Ricardo Martin-Brualla, and Pratul~P. Srinivasan.
        \newblock Mip-nerf: A multiscale representation for anti-aliasing neural radiance fields.
        \newblock \emph{2021 IEEE/CVF International Conference on Computer Vision (ICCV)}, pages 5835--5844, 2021.
        
        \bibitem[Barron et~al.(2022)Barron, Mildenhall, Verbin, Srinivasan, and Hedman]{barron2022mipnerf360}
        Jonathan~T. Barron, Ben Mildenhall, Dor Verbin, Pratul~P. Srinivasan, and Peter Hedman.
        \newblock Mip-nerf 360: Unbounded anti-aliased neural radiance fields.
        \newblock \emph{CVPR}, 2022.
        
        \bibitem[Cao and Johnson(2023)]{HexPlane}
        Ang Cao and Justin Johnson.
        \newblock Hexplane: A fast representation for dynamic scenes.
        \newblock \emph{2023 IEEE/CVF Conference on Computer Vision and Pattern Recognition (CVPR)}, pages 130--141, 2023.
        
        \bibitem[Cao et~al.(2021)Cao, Radosavovic, Kanazawa, and Malik]{cao2021reconstructingRHO}
        Zhe Cao, Ilija Radosavovic, Angjoo Kanazawa, and Jitendra Malik.
        \newblock Reconstructing hand-object interactions in the wild.
        \newblock In \emph{Proceedings of the IEEE/CVF International Conference on Computer Vision}, pages 12417--12426, 2021.
        
        \bibitem[Chen et~al.(2022)Chen, Xu, Geiger, Yu, and Su]{TensoRF}
        Anpei Chen, Zexiang Xu, Andreas Geiger, Jingyi Yu, and Hao Su.
        \newblock Tensorf: Tensorial radiance fields.
        \newblock In \emph{European Conference on Computer Vision (ECCV)}, 2022.
        
        \bibitem[Chen et~al.(2023)Chen, Funkhouser, Hedman, and Tagliasacchi]{mobile-nerf}
        Zhiqin Chen, Thomas Funkhouser, Peter Hedman, and Andrea Tagliasacchi.
        \newblock Mobilenerf: Exploiting the polygon rasterization pipeline for efficient neural field rendering on mobile architectures.
        \newblock In \emph{The Conference on Computer Vision and Pattern Recognition (CVPR)}, 2023.
        
        \bibitem[Cheng et~al.(2023)Cheng, Shan, Hassen, Higgins, and Fouhey]{cheng2023towards}
        Tianyi Cheng, Dandan Shan, Ayda Hassen, Richard Higgins, and David Fouhey.
        \newblock Towards a richer 2d understanding of hands at scale.
        \newblock \emph{Advances in Neural Information Processing Systems}, 36:\penalty0 30453--30465, 2023.
        
        \bibitem[Corona et~al.(2020)Corona, Pumarola, Alenya, Moreno-Noguer, and Rogez]{corona2020ganhand}
        Enric Corona, Albert Pumarola, Guillem Alenya, Francesc Moreno-Noguer, and Gr{\'e}gory Rogez.
        \newblock Ganhand: Predicting human grasp affordances in multi-object scenes.
        \newblock In \emph{Proceedings of the IEEE/CVF conference on computer vision and pattern recognition}, pages 5031--5041, 2020.
        
        \bibitem[Damen et~al.(2018)Damen, Doughty, Farinella, Fidler, Furnari, Kazakos, Moltisanti, Munro, Perrett, Price, and Wray]{Damen2018EPICKITCHENS}
        Dima Damen, Hazel Doughty, Giovanni~Maria Farinella, Sanja Fidler, Antonino Furnari, Evangelos Kazakos, Davide Moltisanti, Jonathan Munro, Toby Perrett, Will Price, and Michael Wray.
        \newblock Scaling egocentric vision: The epic-kitchens dataset.
        \newblock In \emph{European Conference on Computer Vision (ECCV)}, 2018.
        
        \bibitem[Damen et~al.(2022)Damen, Doughty, Farinella, Furnari, Ma, Kazakos, Moltisanti, Munro, Perrett, Price, and Wray]{Damen2022RESCALING}
        Dima Damen, Hazel Doughty, Giovanni~Maria Farinella, Antonino Furnari, Jian Ma, Evangelos Kazakos, Davide Moltisanti, Jonathan Munro, Toby Perrett, Will Price, and Michael Wray.
        \newblock Rescaling egocentric vision: Collection, pipeline and challenges for epic-kitchens-100.
        \newblock \emph{International Journal of Computer Vision (IJCV)}, 130:\penalty0 33–55, 2022.
        
        \bibitem[Darkhalil et~al.(2022)Darkhalil, Shan, Zhu, Ma, Kar, Higgins, Fidler, Fouhey, and Damen]{darkhalil2022epicvisor}
        Ahmad Darkhalil, Dandan Shan, Bin Zhu, Jian Ma, Amlan Kar, Richard Higgins, Sanja Fidler, David Fouhey, and Dima Damen.
        \newblock Epic-kitchens visor benchmark: Video segmentations and object relations.
        \newblock \emph{Advances in Neural Information Processing Systems}, 35:\penalty0 13745--13758, 2022.
        
        \bibitem[Deng et~al.(2022)Deng, Liu, Zhu, and Ramanan]{DS-NeRF}
        Kangle Deng, Andrew Liu, Jun-Yan Zhu, and Deva Ramanan.
        \newblock Depth-supervised {NeRF}: Fewer views and faster training for free.
        \newblock In \emph{Proceedings of the IEEE/CVF Conference on Computer Vision and Pattern Recognition (CVPR)}, 2022.
        
        \bibitem[Du et~al.(2020)Du, Zhang, Yu, Tenenbaum, and Wu]{df_nerf_4}
        Yilun Du, Yinan Zhang, Hong-Xing Yu, Joshua~B. Tenenbaum, and Jiajun Wu.
        \newblock Neural radiance flow for 4d view synthesis and video processing.
        \newblock \emph{2021 IEEE/CVF International Conference on Computer Vision (ICCV)}, pages 14304--14314, 2020.
        
        \bibitem[Fan et~al.(2024)Fan, Parelli, Kadoglou, Kocabas, Chen, Black, and Hilliges]{fan2024hold}
        Zicong Fan, Maria Parelli, Maria~Eleni Kadoglou, Muhammed Kocabas, Xu Chen, Michael~J Black, and Otmar Hilliges.
        \newblock {HOLD}: Category-agnostic 3d reconstruction of interacting hands and objects from video.
        \newblock 2024.
        
        \bibitem[Fang et~al.(2022)Fang, Yi, Wang, Xie, Zhang, Liu, Nie\ss{}ner, and Tian]{TiNeuVox}
        Jiemin Fang, Taoran Yi, Xinggang Wang, Lingxi Xie, Xiaopeng Zhang, Wenyu Liu, Matthias Nie\ss{}ner, and Qi Tian.
        \newblock Fast dynamic radiance fields with time-aware neural voxels.
        \newblock In \emph{SIGGRAPH Asia 2022 Conference Papers}, 2022.
        
        \bibitem[{Fridovich-Keil and Yu} et~al.(2022){Fridovich-Keil and Yu}, Tancik, Chen, Recht, and Kanazawa]{Plenoxels}
        {Fridovich-Keil and Yu}, Matthew Tancik, Qinhong Chen, Benjamin Recht, and Angjoo Kanazawa.
        \newblock Plenoxels: Radiance fields without neural networks.
        \newblock In \emph{CVPR}, 2022.
        
        \bibitem[Grauman et~al.(2022)Grauman, Westbury, Byrne, Chavis, Furnari, Girdhar, Hamburger, Jiang, Liu, Liu, et~al.]{grauman2022ego4d}
        Kristen Grauman, Andrew Westbury, Eugene Byrne, Zachary Chavis, Antonino Furnari, Rohit Girdhar, Jackson Hamburger, Hao Jiang, Miao Liu, Xingyu Liu, et~al.
        \newblock Ego4d: Around the world in 3,000 hours of egocentric video.
        \newblock In \emph{Proceedings of the IEEE/CVF Conference on Computer Vision and Pattern Recognition}, pages 18995--19012, 2022.
        
        \bibitem[Grauman et~al.(2023)Grauman, Westbury, Torresani, Kitani, Malik, Afouras, Ashutosh, Baiyya, Bansal, Boote, Byrne, Chavis, Chen, Cheng, Chu, Crane, Dasgupta, Dong, Escobar, Forigua, Gebreselasie, Haresh, Huang, Islam, Jain, Khirodkar, Kukreja, Liang, Liu, Majumder, Mao, Martin, Mavroudi, Nagarajan, Ragusa, Ramakrishnan, Seminara, Somayazulu, Song, Su, Xue, Zhang, Zhang, Castillo, Chen, Fu, Furuta, Gonzalez, Gupta, Hu, Huang, Huang, Khoo, Kumar, Kuo, Lakhavani, Liu, Luo, Luo, Meredith, Miller, Oguntola, Pan, Peng, Pramanick, Ramazanova, Ryan, Shan, Somasundaram, Song, Southerland, Tateno, Wang, Wang, Yagi, Yan, Yang, Yu, Zha, Zhao, Zhao, Zhu, Zhuo, Arbelaez, Bertasius, Crandall, Damen, Engel, Farinella, Furnari, Ghanem, Hoffman, Jawahar, Newcombe, Park, Rehg, Sato, Savva, Shi, Shou, and Wray]{egoexo4d}
        Kristen Grauman, Andrew Westbury, Lorenzo Torresani, Kris Kitani, Jitendra Malik, Triantafyllos Afouras, Kumar Ashutosh, Vijay Baiyya, Siddhant Bansal, Bikram Boote, Eugene Byrne, Zach Chavis, Joya Chen, Feng Cheng, Fu-Jen Chu, Sean Crane, Avijit Dasgupta, Jing Dong, Maria Escobar, Cristhian Forigua, Abrham Gebreselasie, Sanjay Haresh, Jing Huang, Md~Mohaiminul Islam, Suyog Jain, Rawal Khirodkar, Devansh Kukreja, Kevin~J Liang, Jia-Wei Liu, Sagnik Majumder, Yongsen Mao, Miguel Martin, Effrosyni Mavroudi, Tushar Nagarajan, Francesco Ragusa, Santhosh~Kumar Ramakrishnan, Luigi Seminara, Arjun Somayazulu, Yale Song, Shan Su, Zihui Xue, Edward Zhang, Jinxu Zhang, Angela Castillo, Changan Chen, Xinzhu Fu, Ryosuke Furuta, Cristina Gonzalez, Prince Gupta, Jiabo Hu, Yifei Huang, Yiming Huang, Weslie Khoo, Anush Kumar, Robert Kuo, Sach Lakhavani, Miao Liu, Mi Luo, Zhengyi Luo, Brighid Meredith, Austin Miller, Oluwatumininu Oguntola, Xiaqing Pan, Penny Peng, Shraman Pramanick, Merey Ramazanova, Fiona Ryan, Wei Shan,
          Kiran Somasundaram, Chenan Song, Audrey Southerland, Masatoshi Tateno, Huiyu Wang, Yuchen Wang, Takuma Yagi, Mingfei Yan, Xitong Yang, Zecheng Yu, Shengxin~Cindy Zha, Chen Zhao, Ziwei Zhao, Zhifan Zhu, Jeff Zhuo, Pablo Arbelaez, Gedas Bertasius, David Crandall, Dima Damen, Jakob Engel, Giovanni~Maria Farinella, Antonino Furnari, Bernard Ghanem, Judy Hoffman, C.~V. Jawahar, Richard Newcombe, Hyun~Soo Park, James~M. Rehg, Yoichi Sato, Manolis Savva, Jianbo Shi, Mike~Zheng Shou, and Michael Wray.
        \newblock Ego-exo4d: Understanding skilled human activity from first- and third-person perspectives, 2023.
        
        \bibitem[Gu et~al.(2024)Gu, Lv, Frost, Green, Straub, and Sweeney]{gu2024egolifter}
        Qiao Gu, Zhaoyang Lv, Duncan Frost, Simon Green, Julian Straub, and Chris Sweeney.
        \newblock Egolifter: Open-world 3d segmentation for egocentric perception.
        \newblock \emph{arXiv preprint arXiv:2403.18118}, 2024.
        
        \bibitem[Gu{\'e}don and Lepetit(2023)]{guedon2023sugar}
        Antoine Gu{\'e}don and Vincent Lepetit.
        \newblock Sugar: Surface-aligned gaussian splatting for efficient 3d mesh reconstruction and high-quality mesh rendering.
        \newblock \emph{arXiv preprint arXiv:2311.12775}, 2023.
        
        \bibitem[Guo et~al.(2021)Guo, Kang, Bao, He, and Zhang]{NeRFReN}
        Yuan{-}Chen Guo, Di Kang, Linchao Bao, Yu He, and Song{-}Hai Zhang.
        \newblock Nerfren: Neural radiance fields with reflections.
        \newblock \emph{CoRR}, abs/2111.15234, 2021.
        
        \bibitem[Guzov et~al.(2022)Guzov, Chibane, Marin, He, Sattler, and Pons-Moll]{guzov2022interactionreplica}
        Vladimir Guzov, Julian Chibane, Riccardo Marin, Yannan He, Torsten Sattler, and Gerard Pons-Moll.
        \newblock Interaction replica: Tracking human-object interaction and scene changes from human motion.
        \newblock \emph{arXiv preprint arXiv:2205.02830}, 2022.
        
        \bibitem[Hassan et~al.(2021)Hassan, Ghosh, Tesch, Tzionas, and Black]{hassan2021populating}
        Mohamed Hassan, Partha Ghosh, Joachim Tesch, Dimitrios Tzionas, and Michael~J Black.
        \newblock Populating 3d scenes by learning human-scene interaction.
        \newblock In \emph{Proceedings of the IEEE/CVF Conference on Computer Vision and Pattern Recognition}, pages 14708--14718, 2021.
        
        \bibitem[Hedman et~al.(2021)Hedman, Srinivasan, Mildenhall, Barron, and Debevec]{baked_nerf_1}
        Peter Hedman, Pratul~P. Srinivasan, Ben Mildenhall, Jonathan~T. Barron, and Paul Debevec.
        \newblock Baking neural radiance fields for real-time view synthesis.
        \newblock \emph{ICCV}, 2021.
        
        \bibitem[Huang et~al.(2022)Huang, Yi, H\"oschle, Safroshkin, Alexiadis, Polikovsky, Scharstein, and Black]{Huang_2022_CVPR}
        Chun-Hao~P. Huang, Hongwei Yi, Markus H\"oschle, Matvey Safroshkin, Tsvetelina Alexiadis, Senya Polikovsky, Daniel Scharstein, and Michael~J. Black.
        \newblock Capturing and inferring dense full-body human-scene contact.
        \newblock In \emph{Proceedings of the IEEE/CVF Conference on Computer Vision and Pattern Recognition (CVPR)}, pages 13274--13285, 2022.
        
        \bibitem[Kerbl et~al.(2023)Kerbl, Kopanas, Leimkuehler, and Drettakis]{original_3dgs}
        Bernhard Kerbl, Georgios Kopanas, Thomas Leimkuehler, and George Drettakis.
        \newblock 3d gaussian splatting for real-time radiance field rendering.
        \newblock \emph{ACM Trans. Graph.}, 42\penalty0 (4), 2023.
        
        \bibitem[Kirillov et~al.(2023)Kirillov, Mintun, Ravi, Mao, Rolland, Gustafson, Xiao, Whitehead, Berg, Lo, et~al.]{kirillov2023segmentSAM}
        Alexander Kirillov, Eric Mintun, Nikhila Ravi, Hanzi Mao, Chloe Rolland, Laura Gustafson, Tete Xiao, Spencer Whitehead, Alexander~C Berg, Wan-Yen Lo, et~al.
        \newblock Segment anything.
        \newblock In \emph{Proceedings of the IEEE/CVF International Conference on Computer Vision}, pages 4015--4026, 2023.
        
        \bibitem[Kirschstein et~al.(2023)Kirschstein, Qian, Giebenhain, Walter, and Nie\ss{}ner]{NeRFSemble}
        Tobias Kirschstein, Shenhan Qian, Simon Giebenhain, Tim Walter, and Matthias Nie\ss{}ner.
        \newblock Nersemble: Multi-view radiance field reconstruction of human heads.
        \newblock \emph{ACM Trans. Graph.}, 42\penalty0 (4), 2023.
        
        \bibitem[Lin et~al.(2016)Lin, Guo, Shao, Jiang, Zhu, and Zhu]{lin2016virtual}
        Jenny Lin, Xingwen Guo, Jingyu Shao, Chenfanfu Jiang, Yixin Zhu, and Song-Chun Zhu.
        \newblock A virtual reality platform for dynamic human-scene interaction.
        \newblock In \emph{SIGGRAPH ASIA 2016 virtual reality meets physical reality: Modelling and simulating virtual humans and environments}, pages 1--4. 2016.
        
        \bibitem[Liu et~al.(2021)Liu, Jiang, Xu, Liu, and Wang]{liu2021semi}
        Shaowei Liu, Hanwen Jiang, Jiarui Xu, Sifei Liu, and Xiaolong Wang.
        \newblock Semi-supervised 3d hand-object poses estimation with interactions in time.
        \newblock In \emph{Proceedings of the IEEE/CVF Conference on Computer Vision and Pattern Recognition}, pages 14687--14697, 2021.
        
        \bibitem[Liu et~al.(2022)Liu, Liu, Jiang, Lyu, Wan, Shen, Liang, Fu, Wang, and Yi]{HOI4D}
        Yunze Liu, Yun Liu, Che Jiang, Kangbo Lyu, Weikang Wan, Hao Shen, Boqiang Liang, Zhoujie Fu, He Wang, and Li Yi.
        \newblock Hoi4d: A 4d egocentric dataset for category-level human-object interaction.
        \newblock In \emph{Proceedings of the IEEE/CVF Conference on Computer Vision and Pattern Recognition (CVPR)}, pages 21013--21022, 2022.
        
        \bibitem[Liu et~al.(2023)Liu, Gao, Meuleman, Tseng, Saraf, Kim, Chuang, Kopf, and Huang]{RDRF}
        Yu-Lun Liu, Chen Gao, Andreas Meuleman, Hung-Yu Tseng, Ayush Saraf, Changil Kim, Yung-Yu Chuang, Johannes Kopf, and Jia-Bin Huang.
        \newblock Robust dynamic radiance fields.
        \newblock In \emph{Proceedings of the IEEE/CVF Conference on Computer Vision and Pattern Recognition}, 2023.
        
        \bibitem[Lu et~al.(2024)Lu, Guo, Hui, Chen, Yang, Tang, Zhu, and Dai]{3DDeformableGaussian}
        Zhicheng Lu, Xiang Guo, Le Hui, Tianrui Chen, Ming Yang, Xiao Tang, Feng Zhu, and Yuchao Dai.
        \newblock 3d geometry-aware deformable gaussian splatting for dynamic view synthesis.
        \newblock In \emph{Proceedings of the IEEE/CVF Conference on Computer Vision and Pattern Recognition}, 2024.
        
        \bibitem[Luiten et~al.(2024)Luiten, Kopanas, Leibe, and Ramanan]{Dynamic3DGaussians}
        Jonathon Luiten, Georgios Kopanas, Bastian Leibe, and Deva Ramanan.
        \newblock Dynamic 3d gaussians: Tracking by persistent dynamic view synthesis.
        \newblock In \emph{3DV}, 2024.
        
        \bibitem[Lv et~al.(2024)Lv, Charron, Moulon, Gamino, Peng, Sweeney, Miller, Tang, Meissner, Dong, et~al.]{lv2024aria}
        Zhaoyang Lv, Nickolas Charron, Pierre Moulon, Alexander Gamino, Cheng Peng, Chris Sweeney, Edward Miller, Huixuan Tang, Jeff Meissner, Jing Dong, et~al.
        \newblock Aria everyday activities dataset.
        \newblock \emph{arXiv preprint arXiv:2402.13349}, 2024.
        
        \bibitem[Ma et~al.(2021)Ma, Li, Liao, Zhang, Wang, Wang, and Sander]{deblur-nerf}
        Li Ma, Xiaoyu Li, Jing Liao, Qi Zhang, Xuan Wang, Jue Wang, and Pedro~V. Sander.
        \newblock Deblur-nerf: Neural radiance fields from blurry images.
        \newblock \emph{2022 IEEE/CVF Conference on Computer Vision and Pattern Recognition (CVPR)}, pages 12851--12860, 2021.
        
        \bibitem[Martin-Brualla et~al.(2021)Martin-Brualla, Radwan, Sajjadi, Barron, Dosovitskiy, and Duckworth]{nerf-in-the-wild}
        Ricardo Martin-Brualla, Noha Radwan, Mehdi S.~M. Sajjadi, Jonathan~T. Barron, Alexey Dosovitskiy, and Daniel Duckworth.
        \newblock {NeRF in the Wild: Neural Radiance Fields for Unconstrained Photo Collections}.
        \newblock In \emph{CVPR}, 2021.
        
        \bibitem[Mildenhall et~al.(2020)Mildenhall, Srinivasan, Tancik, Barron, Ramamoorthi, and Ng]{original_nerf}
        Ben Mildenhall, Pratul~P. Srinivasan, Matthew Tancik, Jonathan~T. Barron, Ravi Ramamoorthi, and Ren Ng.
        \newblock Nerf: Representing scenes as neural radiance fields for view synthesis.
        \newblock In \emph{ECCV}, 2020.
        
        \bibitem[Mildenhall et~al.(2022)Mildenhall, Hedman, Martin-Brualla, Srinivasan, and Barron]{raw-nerf}
        Ben Mildenhall, Peter Hedman, Ricardo Martin-Brualla, Pratul~P. Srinivasan, and Jonathan~T. Barron.
        \newblock {NeRF} in the dark: High dynamic range view synthesis from noisy raw images.
        \newblock \emph{CVPR}, 2022.
        
        \bibitem[M\"uller et~al.(2022)M\"uller, Evans, Schied, and Keller]{instant-ngp}
        Thomas M\"uller, Alex Evans, Christoph Schied, and Alexander Keller.
        \newblock Instant neural graphics primitives with a multiresolution hash encoding.
        \newblock \emph{ACM Trans. Graph.}, 41\penalty0 (4):\penalty0 102:1--102:15, 2022.
        
        \bibitem[Pan et~al.(2023)Pan, Charron, Yang, Peters, Whelan, Kong, Parkhi, Newcombe, and Ren]{pan2023aria}
        Xiaqing Pan, Nicholas Charron, Yongqian Yang, Scott Peters, Thomas Whelan, Chen Kong, Omkar Parkhi, Richard Newcombe, and Yuheng~Carl Ren.
        \newblock Aria digital twin: A new benchmark dataset for egocentric 3d machine perception.
        \newblock In \emph{Proceedings of the IEEE/CVF International Conference on Computer Vision}, pages 20133--20143, 2023.
        
        \bibitem[Park et~al.(2021{\natexlab{a}})Park, Sinha, Barron, Bouaziz, Goldman, Seitz, and Martin-Brualla]{Nerfies}
        Keunhong Park, Utkarsh Sinha, Jonathan~T. Barron, Sofien Bouaziz, Dan~B Goldman, Steven~M. Seitz, and Ricardo Martin-Brualla.
        \newblock Nerfies: Deformable neural radiance fields.
        \newblock \emph{ICCV}, 2021{\natexlab{a}}.
        
        \bibitem[Park et~al.(2021{\natexlab{b}})Park, Sinha, Hedman, Barron, Bouaziz, Goldman, Martin-Brualla, and Seitz]{HyperNeRF}
        Keunhong Park, Utkarsh Sinha, Peter Hedman, Jonathan~T. Barron, Sofien Bouaziz, Dan~B Goldman, Ricardo Martin-Brualla, and Steven~M. Seitz.
        \newblock Hypernerf: A higher-dimensional representation for topologically varying neural radiance fields.
        \newblock \emph{ACM Trans. Graph.}, 40\penalty0 (6), 2021{\natexlab{b}}.
        
        \bibitem[Pumarola et~al.(2020)Pumarola, Corona, Pons-Moll, and Moreno-Noguer]{D-NeRF}
        Albert Pumarola, Enric Corona, Gerard Pons-Moll, and Francesc Moreno-Noguer.
        \newblock D-nerf: Neural radiance fields for dynamic scenes.
        \newblock \emph{2021 IEEE/CVF Conference on Computer Vision and Pattern Recognition (CVPR)}, pages 10313--10322, 2020.
        
        \bibitem[Qu et~al.(2023)Qu, Cui, Zhang, Meng, Ma, Deng, and Wang]{qu2023novel}
        Wentian Qu, Zhaopeng Cui, Yinda Zhang, Chenyu Meng, Cuixia Ma, Xiaoming Deng, and Hongan Wang.
        \newblock Novel-view synthesis and pose estimation for hand-object interaction from sparse views.
        \newblock In \emph{Proceedings of the IEEE/CVF International Conference on Computer Vision}, pages 15100--15111, 2023.
        
        \bibitem[Ren and Gu(2010)]{ren2010figure}
        Xiaofeng Ren and Chunhui Gu.
        \newblock Figure-ground segmentation improves handled object recognition in egocentric video.
        \newblock In \emph{2010 IEEE Computer Society Conference on Computer Vision and Pattern Recognition}, pages 3137--3144. IEEE, 2010.
        
        \bibitem[{Sara Fridovich-Keil and Giacomo Meanti} et~al.(2023){Sara Fridovich-Keil and Giacomo Meanti}, Warburg, Recht, and Kanazawa]{kplanes}
        {Sara Fridovich-Keil and Giacomo Meanti}, Frederik~Rahbæk Warburg, Benjamin Recht, and Angjoo Kanazawa.
        \newblock K-planes: Explicit radiance fields in space, time, and appearance.
        \newblock In \emph{CVPR}, 2023.
        
        \bibitem[Sch\"{o}nberger and Frahm(2016)]{schoenberger2016sfm}
        Johannes~Lutz Sch\"{o}nberger and Jan-Michael Frahm.
        \newblock Structure-from-motion revisited.
        \newblock In \emph{Conference on Computer Vision and Pattern Recognition (CVPR)}, 2016.
        
        \bibitem[Sch\"{o}nberger et~al.(2016)Sch\"{o}nberger, Zheng, Pollefeys, and Frahm]{schoenberger2016mvs}
        Johannes~Lutz Sch\"{o}nberger, Enliang Zheng, Marc Pollefeys, and Jan-Michael Frahm.
        \newblock Pixelwise view selection for unstructured multi-view stereo.
        \newblock In \emph{European Conference on Computer Vision (ECCV)}, 2016.
        
        \bibitem[Schwarz et~al.(2020)Schwarz, Liao, Niemeyer, and Geiger]{GRAF}
        Katja Schwarz, Yiyi Liao, Michael Niemeyer, and Andreas Geiger.
        \newblock Graf: Generative radiance fields for 3d-aware image synthesis.
        \newblock \emph{ArXiv}, abs/2007.02442, 2020.
        
        \bibitem[Shao et~al.(2023)Shao, Zheng, Tu, Liu, Zhang, and Liu]{tensor4d}
        Ruizhi Shao, Zerong Zheng, Hanzhang Tu, Boning Liu, Hongwen Zhang, and Yebin Liu.
        \newblock Tensor4d: Efficient neural 4d decomposition for high-fidelity dynamic reconstruction and rendering.
        \newblock In \emph{Proceedings of the IEEE Conference on Computer Vision and Pattern Recognition}, 2023.
        
        \bibitem[Somasundaram et~al.(2023)Somasundaram, Dong, Tang, Straub, Yan, Goesele, Engel, De~Nardi, and Newcombe]{somasundaram2023project}
        Kiran Somasundaram, Jing Dong, Huixuan Tang, Julian Straub, Mingfei Yan, Michael Goesele, Jakob~Julian Engel, Renzo De~Nardi, and Richard Newcombe.
        \newblock Project aria: A new tool for egocentric multi-modal ai research.
        \newblock \emph{arXiv preprint arXiv:2308.13561}, 2023.
        
        \bibitem[Sun et~al.(2022)Sun, Sun, and Chen]{DVGO}
        Cheng Sun, Min Sun, and Hwann{-}Tzong Chen.
        \newblock Direct voxel grid optimization: Super-fast convergence for radiance fields reconstruction.
        \newblock In \emph{CVPR}, 2022.
        
        \bibitem[Tekin et~al.(2019)Tekin, Bogo, and Pollefeys]{tekin2019h+}
        Bugra Tekin, Federica Bogo, and Marc Pollefeys.
        \newblock H+ o: Unified egocentric recognition of 3d hand-object poses and interactions.
        \newblock In \emph{Proceedings of the IEEE/CVF conference on computer vision and pattern recognition}, pages 4511--4520, 2019.
        
        \bibitem[Tschernezki et~al.(2024)Tschernezki, Darkhalil, Zhu, Fouhey, Laina, Larlus, Damen, and Vedaldi]{epicfields}
        Vadim Tschernezki, Ahmad Darkhalil, Zhifan Zhu, David Fouhey, Iro Laina, Diane Larlus, Dima Damen, and Andrea Vedaldi.
        \newblock Epic fields: Marrying 3d geometry and video understanding, 2024.
        
        \bibitem[Verbin et~al.(2022)Verbin, Hedman, Mildenhall, Zickler, Barron, and Srinivasan]{Ref-NeRF}
        Dor Verbin, Peter Hedman, Ben Mildenhall, Todd Zickler, Jonathan~T. Barron, and Pratul~P. Srinivasan.
        \newblock {Ref-NeRF}: Structured view-dependent appearance for neural radiance fields.
        \newblock \emph{CVPR}, 2022.
        
        \bibitem[Wang et~al.(2022)Wang, Tan, Li, Tian, and Liu]{MixVoxels}
        Feng Wang, Sinan Tan, Xinghang Li, Zeyue Tian, and Huaping Liu.
        \newblock Mixed neural voxels for fast multi-view video synthesis.
        \newblock \emph{2023 IEEE/CVF International Conference on Computer Vision (ICCV)}, pages 19649--19659, 2022.
        
        \bibitem[Wang et~al.(2004)Wang, Bovik, Sheikh, and Simoncelli]{1284395}
        Zhou Wang, A.C. Bovik, H.R. Sheikh, and E.P. Simoncelli.
        \newblock Image quality assessment: from error visibility to structural similarity.
        \newblock \emph{IEEE Transactions on Image Processing}, 13\penalty0 (4):\penalty0 600--612, 2004.
        
        \bibitem[Wen et~al.(2023)Wen, Tremblay, Blukis, Tyree, Muller, Evans, Fox, Kautz, and Birchfield]{wen2023bundlesdf}
        Bowen Wen, Jonathan Tremblay, Valts Blukis, Stephen Tyree, Thomas Muller, Alex Evans, Dieter Fox, Jan Kautz, and Stan Birchfield.
        \newblock Bundlesdf: Neural 6-dof tracking and 3d reconstruction of unknown objects.
        \newblock \emph{CVPR}, 2023.
        
        \bibitem[Wong et~al.(2021)Wong, Li, Nie{\ss}ner, and Mitra]{wong2021rigidfusion}
        Yu-Shiang Wong, Changjian Li, Matthias Nie{\ss}ner, and Niloy~J Mitra.
        \newblock Rigidfusion: Rgb-d scene reconstruction with rigidly-moving objects.
        \newblock In \emph{Computer Graphics Forum}, pages 511--522. Wiley Online Library, 2021.
        
        \bibitem[Wu et~al.(2023)Wu, Yi, Fang, Xie, Zhang, Wei, Liu, Tian, and Xinggang]{4DGaussianSplatting}
        Guanjun Wu, Taoran Yi, Jiemin Fang, Lingxi Xie, Xiaopeng Zhang, Wei Wei, Wenyu Liu, Qi Tian, and Wang Xinggang.
        \newblock 4d gaussian splatting for real-time dynamic scene rendering.
        \newblock \emph{arXiv preprint arXiv:2310.08528}, 2023.
        
        \bibitem[Xian et~al.(2020)Xian, Huang, Kopf, and Kim]{dynamic_nerf_per_timestamp_2}
        Wenqi Xian, Jia-Bin Huang, Johannes Kopf, and Changil Kim.
        \newblock Space-time neural irradiance fields for free-viewpoint video.
        \newblock \emph{2021 IEEE/CVF Conference on Computer Vision and Pattern Recognition (CVPR)}, pages 9416--9426, 2020.
        
        \bibitem[Xu et~al.(2022)Xu, Xu, Philip, Bi, Shu, Sunkavalli, and Neumann]{Point-NeRF}
        Qiangeng Xu, Zexiang Xu, Julien Philip, Sai Bi, Zhixin Shu, Kalyan Sunkavalli, and Ulrich Neumann.
        \newblock Point-nerf: Point-based neural radiance fields.
        \newblock \emph{2022 IEEE/CVF Conference on Computer Vision and Pattern Recognition (CVPR)}, pages 5428--5438, 2022.
        
        \bibitem[Yang et~al.(2023{\natexlab{a}})Yang, Gao, Li, Gao, Wang, and Zheng]{yang2023track}
        Jinyu Yang, Mingqi Gao, Zhe Li, Shang Gao, Fangjing Wang, and Feng Zheng.
        \newblock Track anything: Segment anything meets videos, 2023{\natexlab{a}}.
        
        \bibitem[Yang et~al.(2021)Yang, Zhan, Li, Xu, Li, and Lu]{yang2021cpf}
        Lixin Yang, Xinyu Zhan, Kailin Li, Wenqiang Xu, Jiefeng Li, and Cewu Lu.
        \newblock Cpf: Learning a contact potential field to model the hand-object interaction.
        \newblock In \emph{Proceedings of the IEEE/CVF International Conference on Computer Vision}, pages 11097--11106, 2021.
        
        \bibitem[Yang et~al.(2023{\natexlab{b}})Yang, Gao, Zhou, Jiao, Zhang, and Jin]{Deformable3DGaussians}
        Ziyi Yang, Xinyu Gao, Wen Zhou, Shaohui Jiao, Yuqing Zhang, and Xiaogang Jin.
        \newblock Deformable 3d gaussians for high-fidelity monocular dynamic scene reconstruction.
        \newblock \emph{arXiv preprint arXiv:2309.13101}, 2023{\natexlab{b}}.
        
        \bibitem[Ye et~al.(2023{\natexlab{a}})Ye, Danelljan, Yu, and Ke]{ye2023gaussiangrouping}
        Mingqiao Ye, Martin Danelljan, Fisher Yu, and Lei Ke.
        \newblock Gaussian grouping: Segment and edit anything in 3d scenes.
        \newblock \emph{arXiv preprint arXiv:2312.00732}, 2023{\natexlab{a}}.
        
        \bibitem[Ye et~al.(2022)Ye, Gupta, and Tulsiani]{ye2022hand}
        Yufei Ye, Abhinav Gupta, and Shubham Tulsiani.
        \newblock What's in your hands? 3d reconstruction of generic objects in hands.
        \newblock In \emph{CVPR}, 2022.
        
        \bibitem[Ye et~al.(2023{\natexlab{b}})Ye, Hebbar, Gupta, and Tulsiani]{ye2023vhoi}
        Yufei Ye, Poorvi Hebbar, Abhinav Gupta, and Shubham Tulsiani.
        \newblock Diffusion-guided reconstruction of everyday hand-object interaction clips.
        \newblock In \emph{ICCV}, 2023{\natexlab{b}}.
        
        \bibitem[Yi et~al.(2023)Yi, Huang, Tripathi, Hering, Thies, and Black]{yi2022mime}
        Hongwei Yi, Chun-Hao~P. Huang, Shashank Tripathi, Lea Hering, Justus Thies, and Michael~J. Black.
        \newblock {MIME}: Human-aware {3D} scene generation.
        \newblock In \emph{IEEE Conference on Computer Vision and Pattern Recognition (CVPR)}, pages 12965--12976, 2023.
        
        \bibitem[Yu et~al.(2021)Yu, Li, Tancik, Li, Ng, and Kanazawa]{PlenOctrees}
        Alex Yu, Ruilong Li, Matthew Tancik, Hao Li, Ren Ng, and Angjoo Kanazawa.
        \newblock Plenoctrees for real-time rendering of neural radiance fields.
        \newblock \emph{2021 IEEE/CVF International Conference on Computer Vision (ICCV)}, pages 5732--5741, 2021.
        
        \bibitem[Zhang et~al.(2022)Zhang, Zhou, Stent, and Shi]{zhang2022fineegohos}
        Lingzhi Zhang, Shenghao Zhou, Simon Stent, and Jianbo Shi.
        \newblock Fine-grained egocentric hand-object segmentation: Dataset, model, and applications.
        \newblock In \emph{European Conference on Computer Vision}, pages 127--145. Springer, 2022.
        
        \bibitem[Zhang et~al.(2018)Zhang, Isola, Efros, Shechtman, and Wang]{zhang2018perceptual}
        Richard Zhang, Phillip Isola, Alexei~A Efros, Eli Shechtman, and Oliver Wang.
        \newblock The unreasonable effectiveness of deep features as a perceptual metric.
        \newblock In \emph{CVPR}, 2018.
        
        \bibitem[Zhao et~al.(2022)Zhao, Wang, Zhang, Beeler, and Tang]{zhao2022compositional}
        Kaifeng Zhao, Shaofei Wang, Yan Zhang, Thabo Beeler, and Siyu Tang.
        \newblock Compositional human-scene interaction synthesis with semantic control.
        \newblock In \emph{European Conference on Computer Vision}, pages 311--327. Springer, 2022.
        
        \bibitem[Zhou et~al.(2019)Zhou, Barnes, Jingwan, Jimei, and Hao]{Zhou_2019_CVPR}
        Yi Zhou, Connelly Barnes, Lu Jingwan, Yang Jimei, and Li Hao.
        \newblock On the continuity of rotation representations in neural networks.
        \newblock In \emph{The IEEE Conference on Computer Vision and Pattern Recognition (CVPR)}, 2019.
        
        \bibitem[Zhu and Damen(2023)]{zhu2023get}
        Zhifan Zhu and Dima Damen.
        \newblock Get a grip: Reconstructing hand-object stable grasps in egocentric videos.
        \newblock \emph{arXiv preprint arXiv:2312.15719}, 2023.
        
        \end{thebibliography}

}
\clearpage
\setcounter{page}{1}
\maketitlesupplementary

\section{Static Reconstruction Pipeline}

\begin{figure*}[h]
  \centering
   \includegraphics[width=1.0\linewidth]{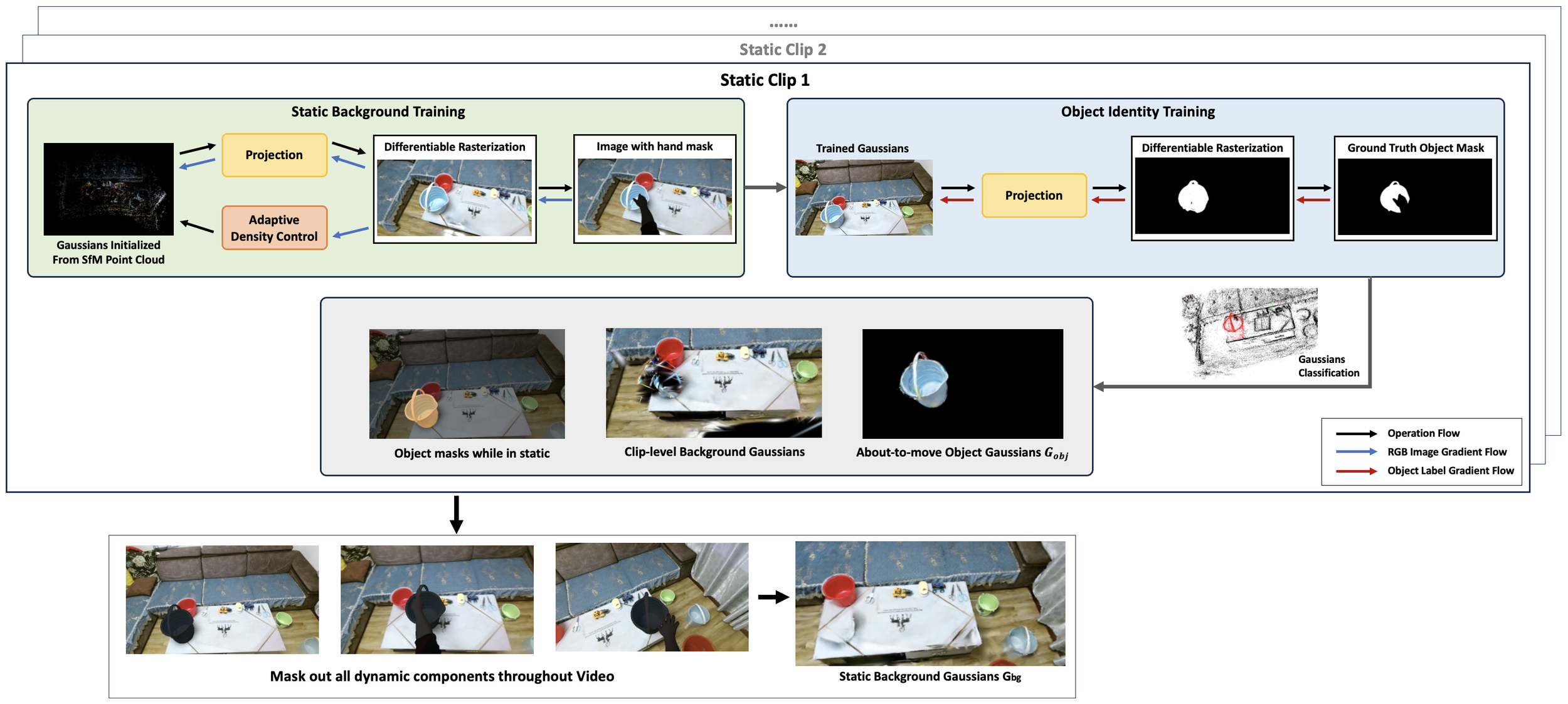}

   \caption{\textbf{Static Reconstruction Pipeline.} We use this pipeline to jointly reconstruct and segment dynamic object from static background from a static egocentric video clip.}
   \label{fig:static_reconstruct}
\end{figure*}

A graphical illustration of our static reconstruction pipeline is provided in \autoref{fig:static_reconstruct}. For each static clip from the egocentric video, we get two Gaussian sets for static background and dynamic object respectively, where the masks projected from object Gaussians are used to complete the previously occluded parts of background.

\section{More Experiment Results}

\subsection{Reconstruction of Dynamic Object}

We also compare our methods with Deformable 3DGS \cite{3DDeformableGaussian} and 4DGS \cite{4DGaussianSplatting} on datasets of frames that exclusively contain the targeted object to exclude the effects from both static background and human body. The datasets are preprocessed to crop out everything except the dynamic object; gradient updates are also disabled on hands that may obscure the object during interaction. 

As illustrated in the supplementary video, the deformation-field-based methods encounter difficulties tracking and modeling the dynamic object\textendash Def-3DGS \cite{Deformable3DGaussians} can only model the very initial movement while 4DGS \cite{4DGaussianSplatting} does not yield any meaningful rendering. Our method is able to accurately track and therefore reconstruct the object with the sequential pose estimation. 

\subsection{Pose Estimation with Larger Time Gap}

In \autoref{tab:ablation_step_size}, we show that we are still able to model the dynamic objects in the scene accurately with a larger time gap in pose estimation. This exchanges a slight performance cost for a significant reduction in training time. We further demonstrate this qualitatively in video results and visualized object trajectories. For simple movements where translation is dominant, the object reconstruction quality and estimated trajectory after pose interpolation remain similar; while for more complicated movements involving rotation, we observe degradation in reconstruction quality

\begin{figure*}[h]
\centering
\setkeys{Gin}{width=0.3\textwidth} %

\begin{minipage}[b]{0.3\linewidth}
    \centering Ours
\end{minipage}%
\begin{minipage}[b]{0.3\linewidth}
    \centering 4DGS
\end{minipage}%
\begin{minipage}[b]{0.3\linewidth}
    \centering Def-3DGS
\end{minipage}%

\smallskip

\begin{minipage}[b]{0.3\linewidth}
    \includegraphics[width=\linewidth]{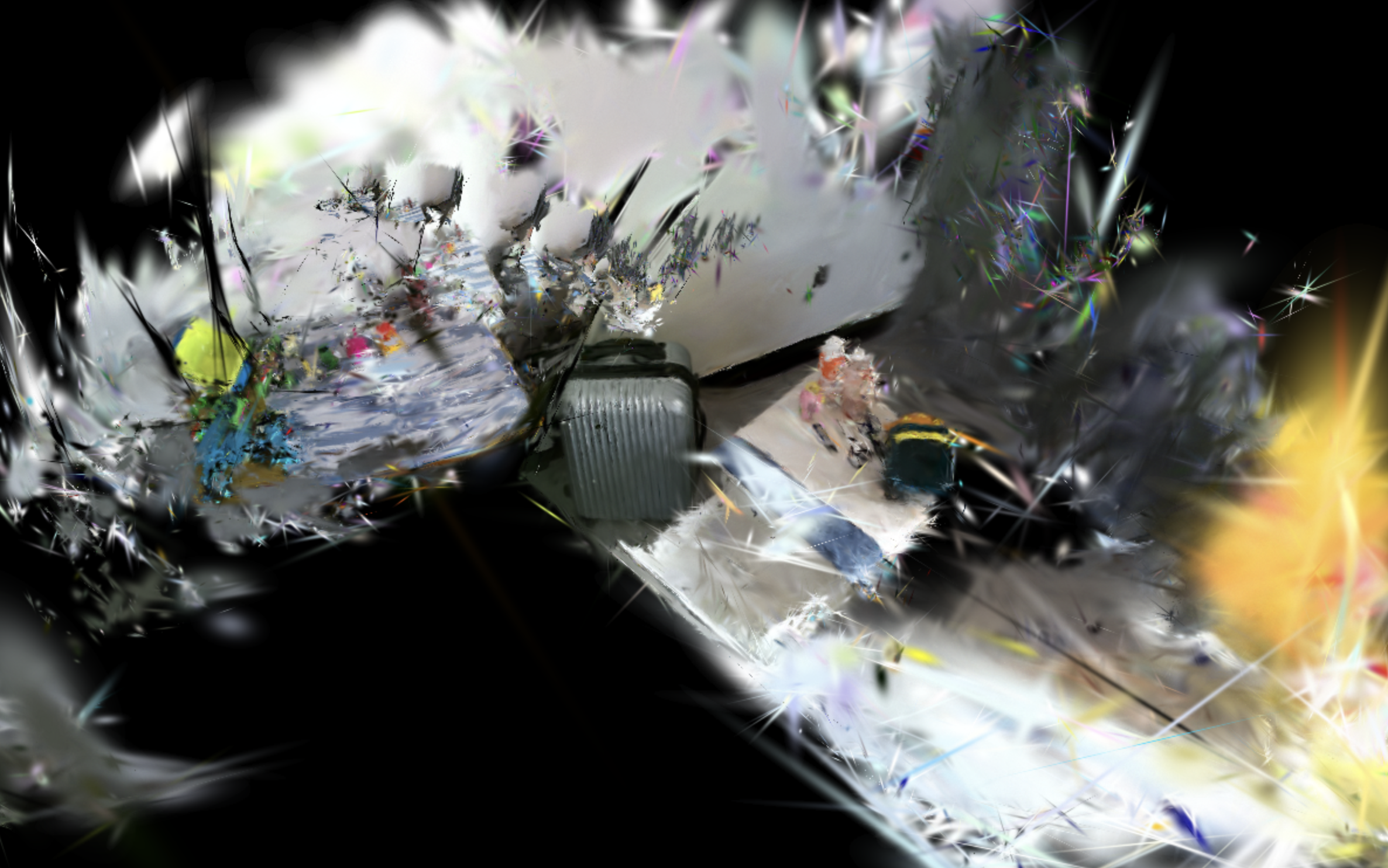}
\end{minipage}%
\begin{minipage}[b]{0.3\linewidth}
    \includegraphics[width=\linewidth]{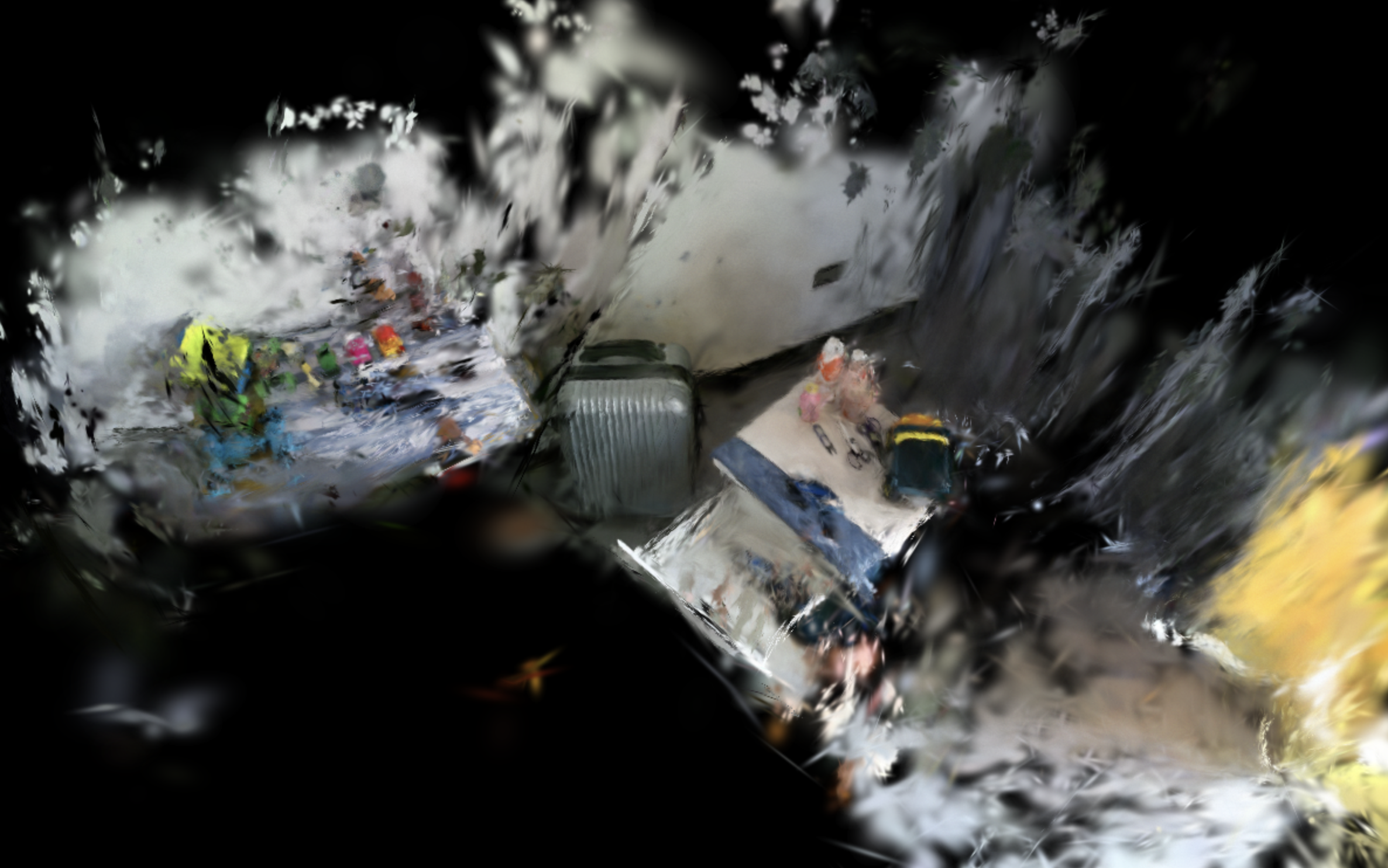}
\end{minipage}%
\begin{minipage}[b]{0.3\linewidth}
    \includegraphics[width=\linewidth]{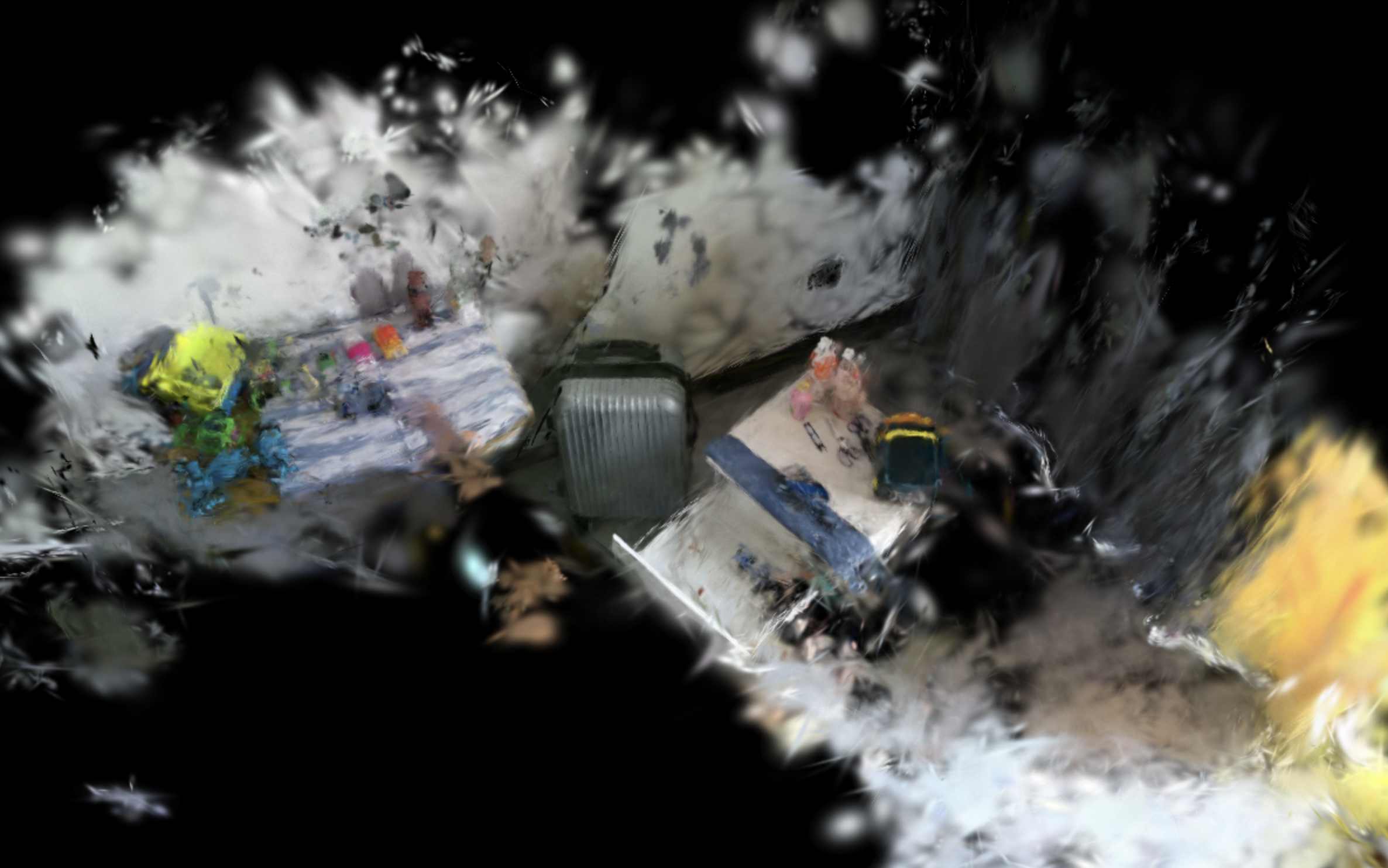}
\end{minipage}%

\caption{\textbf{Comparison with SOTA methods on background reconstruction with larger time gap} All methods suffer from floaters due to sparse training viewpoints}

\label{fig:fail_bg}
\end{figure*}

Although our method can track the dynamic object even across larger time gaps, we cannot properly reconstruct the background, likely due to much sparser information in the presence of strong camera motion. This applies not only to our method but also to both 4DGS and Deformable 3DGS. See examples in \autoref{fig:fail_bg}.

\subsection{Failure Cases}

We mainly observe two types of reconstruction failure. The first type is when the interacted object moves out of sight, which sometimes happens in egocentric video for cameras with a wider field of view, the pose estimation can get stuck in a local minima and lose track of the object in the subsequent frames. The second type arises when the camera registration quality is poor, which is often due to featureless surfaces and insufficient overlap between video frames. The inaccurate camera poses result in a static reconstruction that lacks any geometric understanding of the scene, thereby causing the dynamic reconstruction to fail. 

\end{document}